\documentclass[letterpaper, 10 pt, conference]{ieeeconf}  

\IEEEoverridecommandlockouts                              

\usepackage{amsfonts}
\usepackage{amsmath}
\usepackage{algorithm}
\usepackage{algorithmicx}

\usepackage[noend]{algpseudocode}
\usepackage{mathtools}
\usepackage{array}
\usepackage{graphics}
\usepackage{wrapfig}
\usepackage{subcaption}
\usepackage{float} 
\usepackage{siunitx} 

\usepackage{epsfig}
\usepackage{todonotes}
\usepackage{lipsum}
\usepackage{cite}
\usepackage{todonotes}

\usepackage{lipsum}
\usepackage{makecell}
\usepackage{booktabs}
\usepackage{cleveref}

\DeclareMathAlphabet\mathbfcal{OMS}{cmsy}{b}{n}

\title{\LARGE \bf
Local Advantage Actor-Critic for Robust Multi-Agent \\Deep Reinforcement Learning
}

\author{Yuchen Xiao, Xueguang Lyu, Christopher Amato
\thanks{Khoury College of Computer Sciences, Northeastern University, Boston, MA 02115, USA \{xiao.yuch, lu.xue, c.amato\}@northeastern.edu}
}

\begin{document}

\maketitle
\thispagestyle{empty}
\pagestyle{empty}

\begin{abstract}

    Policy gradient methods have become popular in multi-agent reinforcement learning, but they suffer from high variance due to the presence of environmental stochasticity and exploring agents (i.e., non-stationarity), which is potentially worsened by the difficulty in credit assignment. 
    As a result, there is a need for a method that is not only capable of efficiently solving the above two problems but also robust enough to solve a variety of tasks. 
    To this end, we propose a new multi-agent policy gradient method, called \emph{Robust Local Advantage} (ROLA) \emph{Actor-Critic}. 
    ROLA allows each agent to learn an individual action-value function as a \emph{local critic} as well as ameliorating environment non-stationarity via a novel centralized training approach based on a \emph{centralized critic}. 
    By using this \emph{local critic}, each agent calculates a baseline to reduce variance on its policy gradient estimation, which results in an expected advantage action-value over other agents' choices that implicitly improves credit assignment. 
    We evaluate ROLA across diverse benchmarks and show its robustness and effectiveness over a number of state-of-the-art multi-agent policy gradient algorithms.   

\end{abstract}

\section{INTRODUCTION}

\emph{Multi-agent reinforcement learning } (MARL) has shown many successes in solving real-world multi-robot tasks, such as hallway navigation~\cite{CoRL20-Park}, autonomous vehicles in merging traffic~\cite{TangICCV2019}, warehouse tool delivery~\cite{xiao_icra_2020} and unmanned aerial vehicle coordination~\cite{AlonMARL}.
Recently, MARL has made significant progress on learning decentralized policies for a group of agents to achieve collaborative behaviors by operating based on only local information under uncertainty~\cite{Lucian_Survey,smac,hideseek,liu2018emergent}. 
In practice, having decentralized policies for each robot is preferable since fast and perfect online communication is often hard to guarantee in many real-world applications.

MARL methods often build off of single-agent RL methods. 
In terms of taking advantage of single-agent reinforcement learning techniques, the \emph{independent learning} (IL) framework is the simplest solution, allowing each agent to learn an individual policy in such environments~\cite{tan1993multi}.
Although IL may sometimes work in practice, it encounters a crucial theoretical issue: the environment becomes non-stationary from each agent's perspective as other agents explore and update policies.
This so-called environmental non-stationarity is known to generate a high variance on value and gradient estimations and impedes agents from collaborating well.

\emph{Centralized training with decentralized execution} (CTDE)~\cite{OliehoekSV08,KraemerB16} has been a very promising learning framework for improving solution quality while maintaining decentralized execution in MARL.
CTDE has been implemented with both value-based~\cite{QMIX,WQMIX,MAVEN,ROMA} and policy-gradient-based approaches~\cite{COMA,MADDPG,M3DDPG,SQDDPG,MAAC,LIIR}.
In particular, policy gradient algorithms based on an \emph{actor-critic} framework have become prominent for implementing the CTDE paradigm. The key idea is to train a \emph{centralized critic} conditioned on accessible global information for directing each decentralized actor's optimization.
This \emph{centralized critic} is favored for its stationary learning targets, overcoming the major theoretical problem in IL, and has become the basis of many recent advances.
However, with a global reward function in multi-agent cooperative problems, simply applying a shared \emph{centralized critic} to compute the gradient for each agent's action still causes a severe \emph{credit assignment} issue. It introduces extra variance on gradient estimation for each agent's policy since the critic conditions on joint observations and actions, where the joint space is of exponential size.    

COMA~\cite{COMA}, as a representative multi-agent actor-critic-based method, achieves variance reduction by using a counterfactual baseline, inspired from \emph{difference rewards}~\cite{tumer-wolpert_acs01}, to credit each agent. 
The counterfactual baseline is a promising idea, but it estimates the contribution of each agent's action by marginalizing over only the corresponding agent's counterfactual action choices while keeping other agents' actions fixed.
As a result, the estimation variance caused by other agents' explorations still exists, which leads to noisy credit assignments. 
SQDDPG~\cite{SQDDPG} extends COMA's idea to capture the average contribution of an agent's action by sequentially adding the agent into a set of sampled coalitions.
Although it theoretically improves the effectiveness of resolving the two problems mentioned above, it has a strong requirement on a prior distribution over the coalition space, which is often not available without having good knowledge of a given domain's properties.  
MAAC~\cite{MAAC} utilizes a centralized attention mechanism to solve tasks that require agents to selectively focus on different things in order to produce rich collaborations. However, it still suffers from inefficient credit assignment and variance reduction due to the pure dependency on COMA's counterfactual scheme.
LIIR~\cite{LIIR} explicitly approximates each agent's individual reward based on global reward signals, and its performance highly depends on the underlying complexity of the ground truth credit assignment shaped by the domain reward function.   

Value function factorization has also become popular in MARL for learning individual Q-functions~\cite{QMIX,WQMIX,MAVEN,ROMA}. 
Such methods have recently been introduced into actor-critic frameworks~\cite{VDAC,DOP} to learn critics. 
However, these factorization methods have an inherent limitation caused by restricting the relationship between the joint Q-values and decentralized Q-values, such as assuming a linear summation constraint, a non-linear monotonic constraint, or other weighted constraints.
These constraints prevent the methods from learning the true joint action-value function in general and potentially limit each method to work well in particular tasks (such as the SMAC domains~\cite{smac} in most related papers) but perform worse in some other domains~\cite{CM3}.

In this paper, we propose a novel policy gradient actor-critic framework that not only is able to efficiently address both credit assignment and learning variance problems but also possesses high robustness, where the `robustness' means that the proposed method always gains outstanding performance over multiple domains. To this end, we introduce a \emph{Robust Local Advantage} (ROLA) \emph{Actor-Critic} policy gradient algorithm to learn decentralized policies. 

ROLA features a new centralized training approach that incorporates a \emph{centralized critic} into each agent's \emph{local critic} optimization with the benefits including: 
a) alleviating the effectiveness of environmental non-stationarity; 
b) counteracting the overestimation often existing in learning with a single value approximator; 
c) generating extra joint-action selections from the centralized perspective as training data rather than using only decentralized trajectories in existing CTDE-based policy gradient approaches. 
d) facilitating agents to avoid local optima to learn good collaborations. 
The \emph{local critic} is also allowed to access extra global information but only conditions on each agent's own action. The resulting critic provides an estimated action-value over other agents' behaviors, implicitly completing the credit assignment. Furthermore, by using this \emph{local critic}, each agent can obtain a low-variance 
advantage action-value estimation that can be used for policy updates.

\section{Background}
\label{sec:bg}

We start by formalizing our fully cooperative multi-agent reinforcement learning problem and then introduce the fundamental reinforcement learning algorithms that our new approach builds on.

\subsection{Dec-POMDPs}

In a fully cooperative task, a team of agents behaves in a decentralized operating manner, individually choosing actions purely based on local observations, with state and outcome uncertainties can be described as a \emph{decentralized partially observable Markov decision process} (Dec-POMDP)~\cite{Oliehoek}, defined as a tuple $\langle I,S,A,\Omega,T,O,R \rangle$. 
$I = \{1,...,n\}$ represents a finite set of agents; 
$S$ is a finite set of environment true states; 
$A = \times_i A_i$ and $\Omega = \times_i \Omega_i$ stand for the joint action space and the joint observation space over agents respectively. 
At every time-step, each agent $i$ synchronously executes its own action $a_i$, which induces that the environment transits from the current state $s$ to a new state $s'$ according to the state transition function $T(s,\vec{a},s') = P(s'| s,\vec{a})$, where a joint action formed as $\vec{a}=\times_ia_i \in A$; 
meanwhile, an immediate reward $r(s, \vec{a})$ shared over agents is generated by the global reward function $R:S\times A\rightarrow \mathbb{R}$; 
successively, due to the partial observability, agents can only obtain a joint observation $\vec{o}=\times_io_i\in\Omega$ under the new state $s'$, drawn according to the observation function $O(\vec{o},\vec{a},s')=P(\vec{o}|\vec{a},s')$, and each agent's policy $\pi_i(a_i| \tau_i)$ is thus modeled as a mapping from individual action-observation history $\tau_i$ to actions. The objective is to optimize a joint policy $\vec{\pi}=\times_i\pi_i$ that maximizes the expectation of a long-term discounted return from an initial state $s_0$, $V^{\vec{\pi}}(s_0)=\mathbb{E}_{\vec{\pi}}\big[\sum_{t=0}^{h-1}\gamma^tr(s_t, \vec{a}_t)| s_0\big]$, where $h$ is the horizon of the task and $\gamma\in[0,1]$ is a discount factor.

\subsection{Advantage Actor-Critic (A2C)}

Policy gradient is one popular reinforcement learning technique with the aim to directly optimize a parameterized policy $\pi_{\theta}$ by performing gradient ascent on the expectation of discounted returns $\mathbb{E}_{\pi_{\theta}}[G]$.
Based  on the single-agent \emph{policy gradient theorem}~\cite{Sutton:1999}, the gradient with respect to policy's parameters in POMDPs can be written as $\nabla_\theta J (\theta)=\mathbb{E}_{\pi_{\theta}}[\nabla_\theta\log\pi_\theta(a|\tau)Q^{\pi_\theta}(a,\tau)]$.
In \emph{actor-critic} framework~\cite{konda2000actor}, the on-policy action-value $Q^{\pi_\theta}$ is approximated by using an action-value function $Q^{\pi_\theta}_\phi$ (critic) learned via \emph{temporal-difference} (TD) learning.
As a common choice, people often train a state-value function in MDPs but history-value function $V^{\pi_\theta}_{\mathbf{w}}(\tau)$ in POMDPs as the critic and incorporate it into the policy gradient in a variance reduction format~\cite{Sutton:1999} ending up with an \emph{advantage actor-critic} (A2C) policy gradient that can be written as:   

\begin{equation}
    \nabla_{\theta}J(\theta) = \mathbb{E}_{\pi_{\theta}}\Big[\nabla_\theta\log\pi_\theta(a|\tau)A(\tau,a)\Big]
    \label{a2c}
\end{equation}
where, $A(\tau,a) = r(\tau, a) + \gamma V^{\pi_\theta}_\mathbf{w}(\tau') - V^{\pi_\theta}_\mathbf{w}(\tau)$.

\subsection{Independent Advantage Actor-Critic (IA2C)}
\label{IA2C}

In the light of \emph{independent Q-learning}~\cite{tan1993multi}, \emph{independent advantage actor-critic} (IA2C)~\cite{COMA} is a straightforward extension of single-agent A2C to multi-agent scenarios.
The specific version of IA2C considered in this paper is formulated as:

\begin{equation}
    \nabla_{\theta_i}J(\theta_i) = \mathbb{E}_{\vec{\pi}_{\theta}}\Big[\nabla_{\theta_i}\log\pi_{\theta_i}(a_i|\tau_i)A(\tau_i,a_i)\Big]
    \label{ia2c}
\end{equation}
where, $A(\tau_i,a_i) = r + \gamma V^{\pi_{\theta_i}}_{\mathbf{w}_i}(\tau_i') - V^{\pi_{\theta_i}}_{\mathbf{w}_i}(\tau_i)$ with a shared reward $r$ assigned by the global reward function $R$. 
Agents independently optimize each own actor $\pi_{\theta_i}$ and critic $V^{\pi_{\theta_i}}_{\mathbf{w}_i}$ purely based on local experiences. 
While decentralized policies can be directly learned in this simple decomposition manner, each independent learner suffers from several innate limitations: 
the difficulty in distinguishing environmental natural stochasticity from other agents' explorations; 
the dilemma of the environmental non-stationarity from a local perspective caused by the existence of other learning agents; 
the tendency of settling at a local optimum due to rare information sharing over agents.

\subsection{Counterfactual Multi-Agent Policy Gradients}

With the mission of addressing the aforementioned limitations in IA2C, \emph{Counterfactual multi-agent} (COMA) policy gradients~\cite{COMA}, as the first A2C-based method adopting \emph{centralized training with decentralized execution} (CTDE), proposes to learn a centralized critic, $ Q_\phi^{\vec{\pi}_\theta}(s,\vec{a})$, conditioning on full state information and all agents' actions to annihilate the non-stationary environment issue, and achieves variance reduction and credit assignment by designing a counterfactual baseline with respect to each agent's policy in an advantage function. The corresponding policy gradient for each agent is then defined as:

\begin{equation}
    \nabla_{\theta_i}J(\theta_i) = \mathbb{E}_{\vec{\pi}_\theta}\biggr[\nabla_{\theta_i}\log\pi_{\theta_i}(a_i|\tau_i)A_i(s, \vec{a})\biggr]
    \label{coma1}
\end{equation}
\begin{equation}
    A_i(s,\vec{a}) = Q_\phi^{\vec{\pi}_\theta}(s,\vec{a}) - \sum_{\dot{a_i}}\pi_{\theta_i}(\dot{a_i}|\tau_i)Q_\phi^{\vec{\pi}_\theta}(s,\dot{a_i},\vec{a}_{-i}) 
    \label{coma2}
\end{equation}

We notice that, in Eq.~\ref{coma2}, the counterfactual baseline is calculated by marginalizing the action of agent $i$ while keeping others' fixed.
However, it is inefficient on both variance reduction and credit assignment due to the freeze of other agents' actions in the counterfactual baseline computation such that:
(a) from agent $i$'s perspective, this counterfactual variance reduction trick will malfunction when the variance is caused by the explorations of other agents rather than agent $i$; 
(b) the credit assignment attained by Eq.\ref{coma2} then becomes just a sampled advantage value for the agent's action $a_i$ under a specific joint-action taken by teammates, which potentially still leads to high variance on the agent's policy gradient related to the action $a_i$ in Eq.~\ref{coma1}. 
It is a variant of the multi-action variance that lives in vanilla multi-agent policy gradients with a single centralized critic~\cite{Lyu_aamas_2021,DOP}.  

The above two underlying flaws in COMA's counterfactual design can also be visualized by taking a further derivation from Eq.~\ref{coma1} as follows:

\begin{align}
    \nabla_{\theta_i}J(\pi_{\theta_i}) & = \mathbb{E}_{\vec{\pi}_\theta}\Big[\nabla_{\theta_i}\log\pi_{\theta_i}(a_i|\tau_i)A_i(s, \vec{a})\Big]\nonumber \\
    & = \mathbb{E}_{\pi_{\theta_i}, \vec{\pi}_{\theta_{-i}}} \Big[\nabla_{\theta_i}\log\pi_{\theta_i}(a_i|\tau_i)A_i(s, a_i, \vec{a}_{-i})\Big] \\ 
    & = \mathbb{E}_{\pi_{\theta_i}}\biggr[\mathbb{E}_{\vec{\pi}_{\theta_{-i}}} \Big[\nabla_{\theta_i}\log\pi_{\theta_i}(a_i|\tau_i)A_i(s, a_i, \vec{a}_{-i})\Big]\biggr]\\ 
    & = \mathbb{E}_{\pi_{\theta_i}}\biggr[\nabla_{\theta_i}\log\pi_{\theta_i}(a_i|\tau_i) \mathbb{E}_{\vec{\pi}_{\theta_{-i}}}\Big[A_i(s, a_i, \vec{a}_{-i})\Big]\biggr] 
    \label{expectA2C}
\end{align}

Eventually, we reach an \emph{expected counterfactual advantage policy gradient} (ECA) as shown in Eq.~\ref{expectA2C}, which can be considered as an extension of the single-agent expected policy gradient~\cite{JMLR:v21:18-012} to multi-agent CTDE framework. 

Eq.~\ref{expectA2C} indicates that, in order to obtain a more accurate value estimation on agent $i$'s action $a_i$ with low variance, we are supposed to either explicitly compute the expected advantage value of $a_i$ or approximate it.
It is a straightforward marginalization process in the centralized training paradigm with a centralized critic, where we can directly compute the expectation using the policies and the critic. 
However, it introduces significant computational costs exponentially increasing with the number of agents and each agent's action space. 
Also, note that obtaining an approximation of ECA is exactly what IA2C is trying to do, but several fatal defects exist as mentioned in section~\ref{IA2C}.    

In this paper, we propose an approach to tackle the above issues by letting each agent learn an individual \emph{local critic} only conditioning on its local action but in a novel on-policy centralized training procedure which relieves the environmental non-stationarity as well as achieving credit assignment implicitly. 
Furthermore, the local critic is then used to calculate a local advantage value for each agent to reduce gradient variance more efficiently.

\section{Approach}

This section presents a new CTDE algorithm called \emph{Robust Local Advantage (ROLA) Actor-Critic}.
The main idea of ROLA is to learn a local action-value function as the critic of each agent but inheriting a centralized insight that facilitates agents to discover good coordinative behaviors. 
ROLA also implicitly carries out multi-agent credit assignment and uses this local critic in advantage action-value estimation to produce variance reduction on each agent's policy gradients.             

Specifically, at the first step, ROLA learns a fully centralized critic, $Q_{\phi}^{\vec{\pi}_\theta}(\mathbf{x}, \vec{a})$. 
As an on-policy evaluation of joint actions, it is updated by minimizing the square TD-error loss:
\begin{equation}
    \mathcal{L}(\phi) = \mathbb{E}_{\vec{\pi}_\theta}\biggr[\big(Q_\phi^{\vec{\pi}_\theta}(\mathbf{x}, \vec{a})-y\big)^2\biggr]
\end{equation}
\begin{equation}
    y = r + \gamma Q_{\phi^-}^{\vec{\pi}_{\theta^-}}(\mathbf{x'}, a_1',...,a_n') \mid _{a_i'\sim\pi_{\theta_i^-}(\tau_i')}
\end{equation}
where, each agent's target policy $\pi_{\theta_i^-}$ with delayed parameters $\theta_i^-$ is used to sample the following action for calculating the target prediction to further stabilizing the learning. 
$\mathbf{x}$ denotes the accessible global signals such as joint observations, joint action-observation histories, or the true environmental state.  

Secondly, ROLA defines a \emph{semi-centralized action-value function}, $Q^{\pi_{\theta_i}}(\mathbf{x}, a_i)$, for each agent $i$, that conditions only on the corresponding agent's action but is allowed to access additional centralized information. 
This implies that $Q^{\pi_{\theta_i}}(\mathbf{x}, a_i)$ satisfies the Bellman equation of either Dec-MDP~\cite{Oliehoek} or Dec-POMDP depending on if the information encoded in $\mathbf{x}$ is the ground truth environment state or action-observation histories. 
We, therefore, are able to use a neural network function approximator with weights $\psi_i$ to estimate $Q^{\pi_{\theta_i}}(\mathbf{x}, a_i)$ via sample-based model-free reinforcement learning approaches. This function approximator is referred to as each agent's \emph{local critic}, represented as $Q_{\psi_i}^{loc}(\mathbf{x}, a_i)$.

Training this \emph{local critic} to provide an accurate value estimation on each agent's action for assisting the decentralized actor optimization via policy gradients is a non-trivial problem, as it does not have access to other agents' actions. 
As a result, it still confronts the curse of the environmental non-stationarity and may diverge the cooperative objective's policies. 
To deal with this issue, we develop a novel \emph{centralized-double-critic} training framework taking advantage of a fully centralized critic to update individual local critics. 

More concretely, in order to let each agent's \emph{local critic}, $Q_{\psi_i}^{loc}(\mathbf{x}, a_i)$, precisely measure the value of a given action $a_i$ under global information $\mathbf{x}$, it has to be capable to capture the expected effect of other agent action selections. 
We achieve this by first using the centralized critic, $Q_{\phi}^{\vec{\pi}_\theta}(\mathbf{x}, \vec{a})$, to sample a joint next action $\vec{a}\,'$ via `softmax' operation, from which the corresponding agent's next action $a_i'$ is then extracted and passed into the target value calculation for computing a local square TD-error being minimized during training:      

\begin{equation}
    \mathcal{L}(\psi_i) = \mathbb{E}_{\vec{\pi}_\theta}\biggr[\big(Q_{\psi_i}^{loc}(\mathbf{x}, a_i) - y\big)^2\biggr]
\end{equation}
\begin{equation}
    y = r + \gamma Q_{\psi_i^-}^{loc}(\mathbf{x}', a_i') \mid _{a_i'\leftarrow \vec{a}\,' \sim \text{softmax} \big(Q_\phi^{\vec{\pi}_{\theta}}(\mathbf{x'},\vec{a}\,')\big)}
    \label{locupdate}
\end{equation}

\begin{figure}[t!]
    \centering
    \includegraphics[height=5cm]{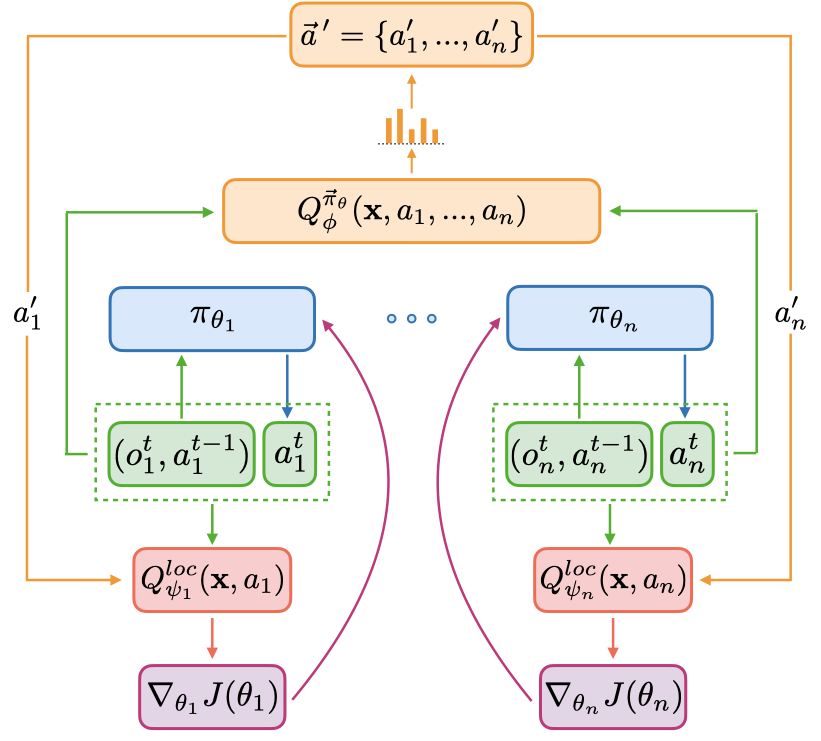}
    \caption{Architecture of ROLA.}
    \label{cola}
\end{figure}

In Eq.~\ref{locupdate}, sampling a joint action from the perspective of the centralized critic incorporates other agents' concurrent behaviors into each agent's \emph{local critic} updating. It endows each agent's \emph{local critic} with the capability of evaluating the utility of the agent's action over teammates' options under a given $\mathbf{x}$, which is how the credit assignment proceeds. 
Furthermore, another strength by applying the sampling in Eq.~\ref{locupdate} is that each agent's \emph{local critic} intrinsically inherits the benefit from a centralized control attitude, being able to facilitate the decentralized actor to be updated towards the cooperative direction even though it only has the agent's own action as input. 
Additionally, the global information considered in the \emph{local critic} does not have to be the same as the one in the centralized critic. There could be many variants in practice depending on the domains' properties. Without loss of generality, we use environment state information in experiments. 

A natural limitation of conventional CTDE-based methods is that the training depends on the actions generated via only decentralized execution. Therefore, it is also important to note that this new idea for learning \emph{local critic} offers centralized action-selection data as extra assets, which breaks through the above limitation of CTDE framework.

In the end, we fuse each agent's \emph{local critic} into the corresponding policy gradient with a local advantage function for each agent's action, which can be expressed as:  

\begin{equation}
    \nabla_{\theta_i} J(\theta_i) = \mathbb{E}_{\vec{\pi}_\theta}\biggr[\nabla_{\theta_i}\log\pi_{\theta_i}(a_i|\tau_i)A_i(\mathbf{x},a_i)\biggr]
\end{equation}

\begin{equation}
    A_i(\mathbf{x}, a_i) = Q_{\psi_i}^{loc}(\mathbf{x}, a_i) - \sum_{\dot{a_i}}\pi_{\theta_i}(\dot{ a_i}|\tau_i)Q_{\psi_i}^{loc}(\mathbf{x},\dot{a_i})
    \label{advlocal}
\end{equation}

Now, to each agent, the expected advantage of taking a particular action under a specific state over whatever other agents behave is assessed in Eq.~\ref{advlocal}, so that the potential punishment caused by other agents can be alleviated. 
Moreover, ROLA calculates a separate baseline using each agent individual \emph{local critic}, serving as a recipe to reduce the variance on policy gradient estimation provoked by the explorations of its own and other agents during training. 

Fig.~\ref{cola} displays the architecture of ROLA. Although we only describe an on-policy version of ROLA above, the off-policy training can be easily done by implementing a replay buffer and importance sampling weights.

\section{Experiments}
\label{exp}

\begin{figure}[b!]
    \centering
    \captionsetup[subfigure]{labelformat=empty}
    \centering
    \subcaptionbox{(a) Capture Target\vspace{2mm}}
        [0.48\linewidth]{\includegraphics[height=2.4cm]{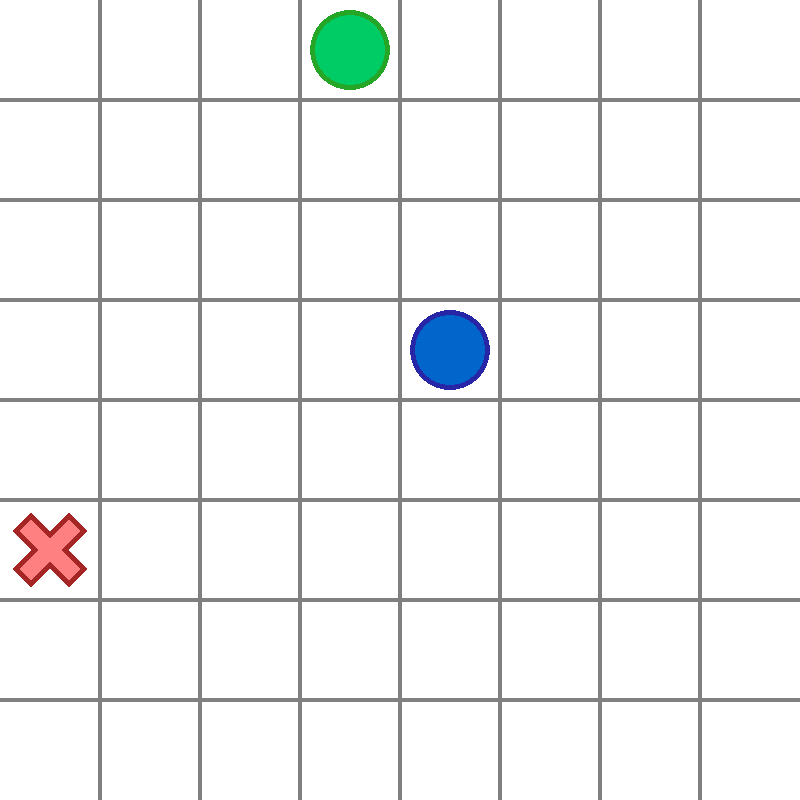}\vspace{1mm}}
    ~
    \centering
    \subcaptionbox{(b) Box Pushing}
        [0.48\linewidth]{\includegraphics[height=2.4cm]{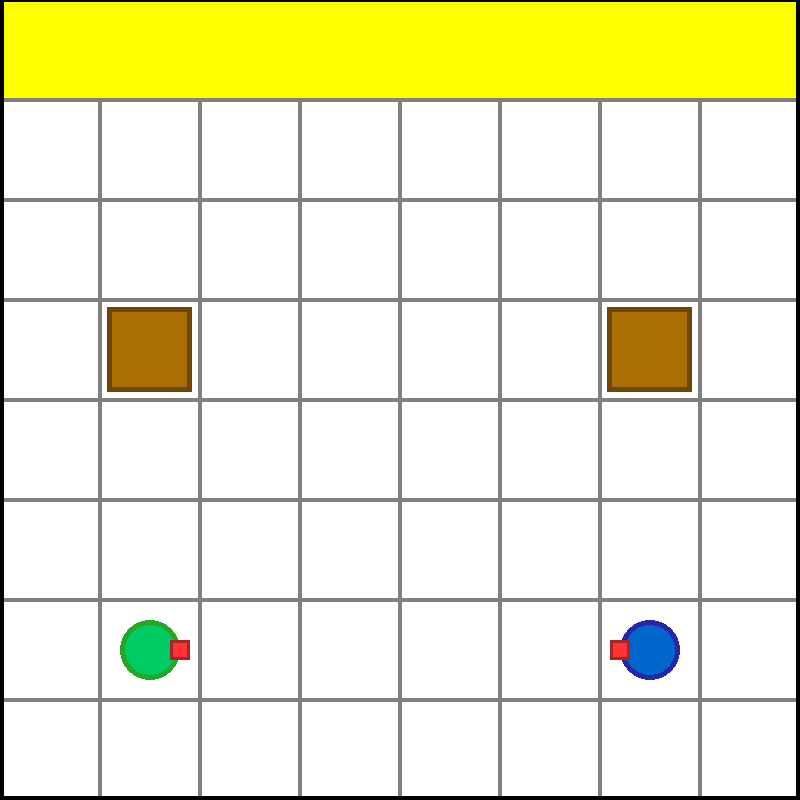}}
    ~
    \centering
    \subcaptionbox{(c) Cooperative Navigation}
        [0.48\linewidth]{\includegraphics[height=2.4cm]{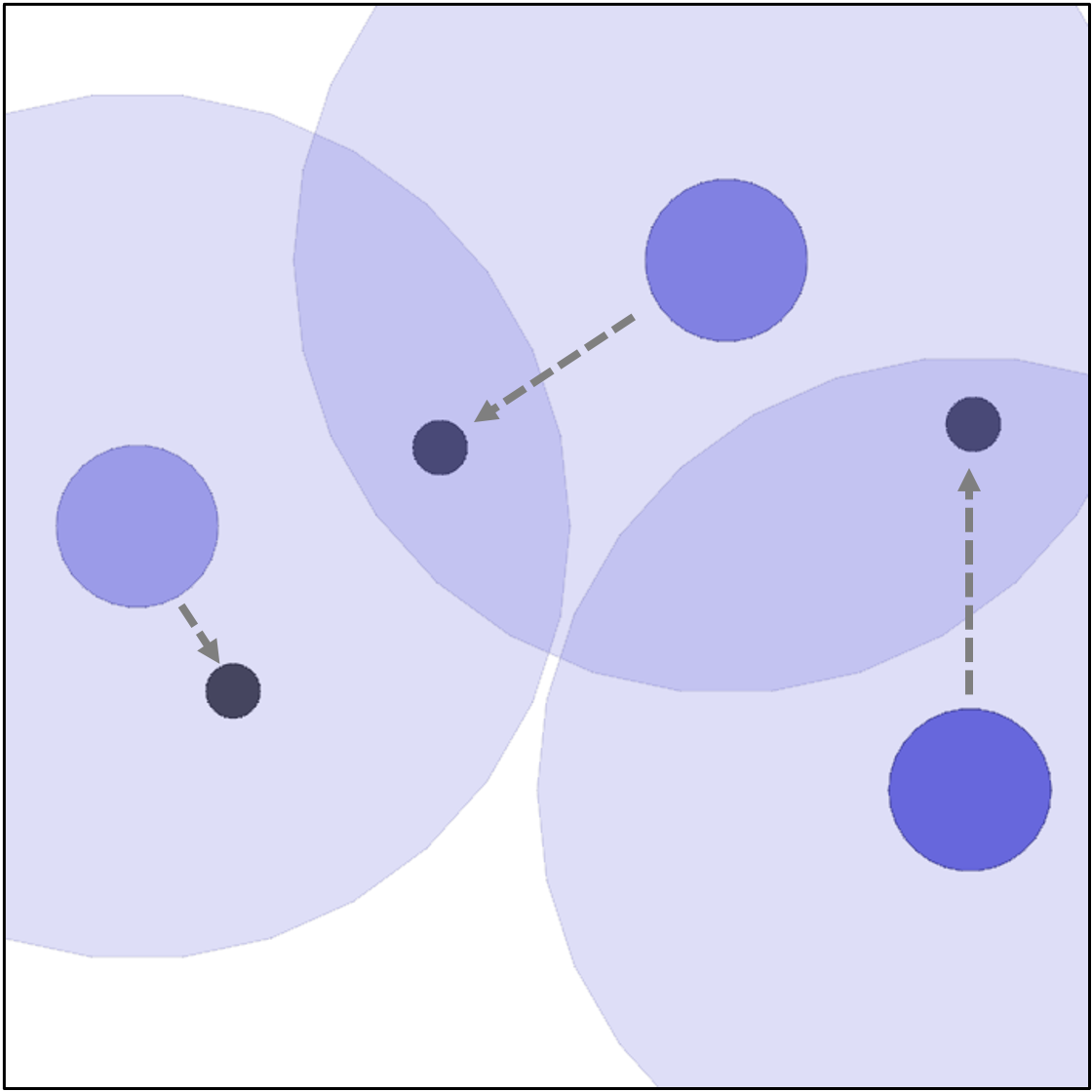}\vspace{1mm}}
    ~
    \centering
    \subcaptionbox{(d) Antipodal Navigation\vspace{2mm}}
        [0.48\linewidth]{\includegraphics[height=2.4cm]{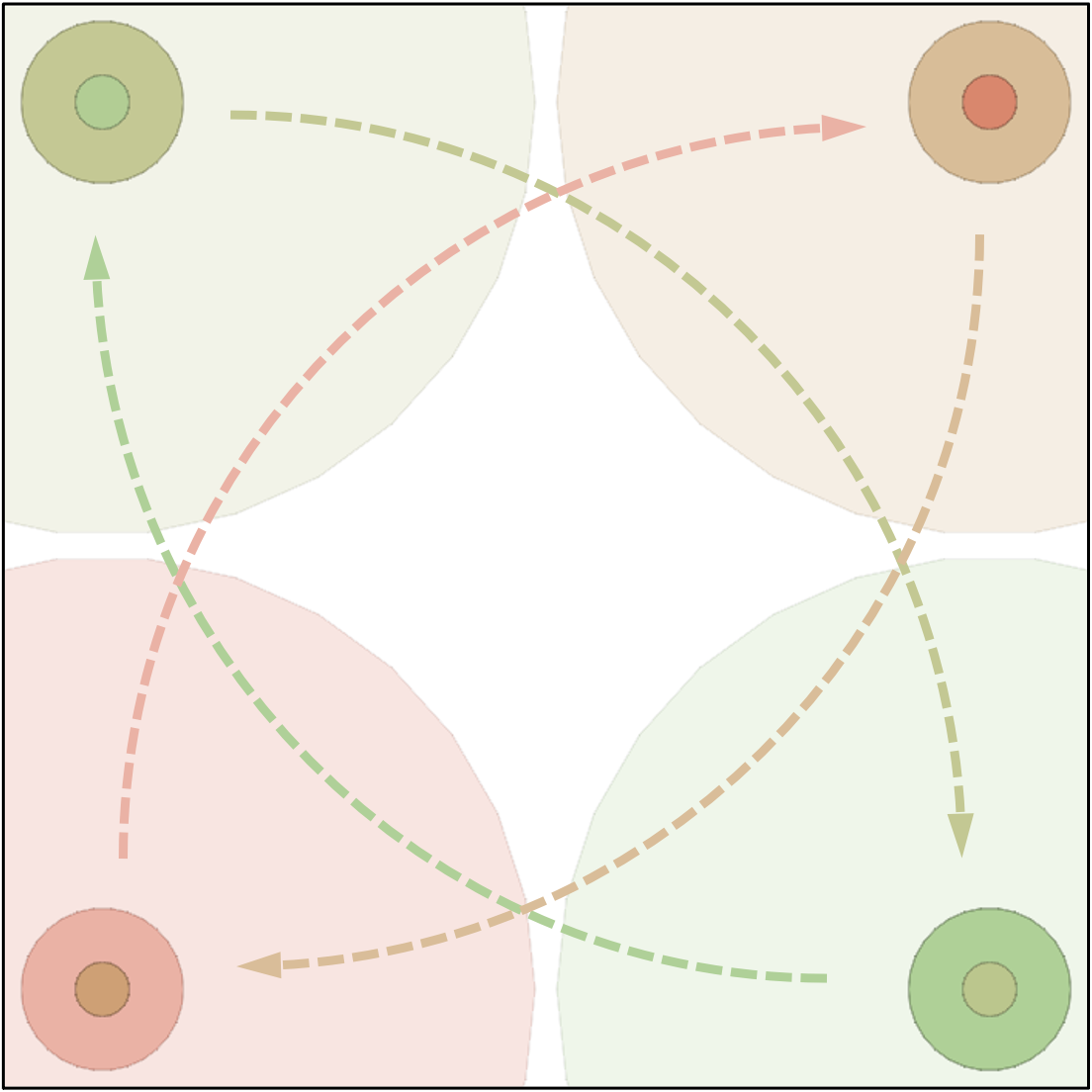}}
    \caption{Experimental domains.}
    \label{domains}
\end{figure}

\begin{table*}[t!]
    \caption {Multi-Agent Actor-Critic Baselines.}
    \centering
    \begin{tabular}{llcccc}
    \toprule
        Algorithm & Feature Description & CTDE & VDN-Based & \makecell{On-Policy \\ Learning} & \makecell{Off-Policy \\ Learning}\\
    \cmidrule(r){1-6}
        IA2C & Refer to Section~\ref{IA2C} & - & - & Yes & -\\
        \\[-1em]
        Central-V~\cite{COMA} & \makecell[l]{Decentralized actors with a state-value function as the centralized critic} & Yes & - & Yes & - \\
        \\[-1em]
        COMA~\cite{COMA} & Counterfactual advantage value & Yes & - & Yes & - \\
        \\[-1em]
        LIIR~\cite{LIIR} & Learn intrinsic reward for each agent & Yes & - & Yes & - \\
        \\[-1em]
        SQDDPG~\cite{SQDDPG}& Approximate marginal contribution of agent's action as local critic  & Yes & - & - & Yes   \\
        & using sampled coalitions & & & \\
        \\[-1em]
        MAAC~\cite{MAAC}  & \makecell[l]{Learn local critics with a centralized attention mechanism} & Yes & - & - & Yes\\
        \\[-1em]
        DOP~\cite{DOP} & Centralized critic is a linear weighted summation of individual critics & Yes & Yes & Yes & Yes \\
        &  optimized by minimizing a mixing of on-policy and off-policy losses & & & \\
        \\[-1em]
        VDAC-mix~\cite{VDAC} & Centralized critic is a non-linear mixing of individual critics & Yes & Yes & Yes & - \\
        \\[-1em]
        VDAC-sum~\cite{VDAC} & Centralized critic is a linear summation of individual critics & Yes & Yes & Yes & - \\
        \\[-1em]
        ECA & Explicitly compute expected counterfactual advantage value & Yes & - & Yes & - \\
    \bottomrule
    \end{tabular} 
    \label{baselines}
\end{table*}

\subsection{Domains Setup}

To investigate the robustness and effectiveness of ROLA versus existing multi-agent policy gradient algorithms, we select four popular benchmark problems: 
Capture Target~\cite{DecHDRQN}, 
a variant version of Box Pushing~\cite{SZuai07}, 
OpenAI Cooperative Navigation~\cite{MADDPG} and Antipodal Navigation~\cite{CM3} with partial observability. The properties of these domains differ from each other in terms of reward density, environmental stochasticity, collaborative format, and the number of agents.

\textbf{Capture Target.} Two agents move around in a toroidal grid world by executing actions: \emph{up}, \emph{down}, \emph{left}, \emph{right} and \emph{stay}, with the transition dynamics as 0.1 probability of arriving at an unintended adjacent cell. 
Only a terminal reward $+1$ can be received when two agents capture the target together by simultaneously locating the cell where the target is. 
Agents can perfectly observe their own positions but cannot receive teammate's location, and they see the target with a probability of 0.7. 
Thus it is a high noisy domain in terms of the environmental stochasticity and the aliased information in the observation, selected for studying the utility of ROLA on variance reduction. 

\textbf{Box Pushing.} The available actions to each agent include: \emph{move forward}, \emph{turn left}, \emph{turn right} and \emph{stay}; and the objective is to push two boxes to the top goal area. 
Each agent can move any box towards the north by executing \emph{move forward} while facing it. 
When any agent successfully pushes one box to the goal, the entire team receives a shared reward $+100$, and the episode also immediately terminates. 
We design this domain on purpose to check if ROLA benefits each agent to quickly deduce its own contribution to a reward, especially when the box is pushed to the goal by the teammate.     

\textbf{Cooperative Navigation.} We extend OpenAI cooperative navigation task with partial observability by imposing a \emph{view range}, as the shallow circle in Fig.~\ref{domains}c, that restricts each agent from capturing the information of other agents and landmarks out of the view. 
Agents have to cooperatively spread out to cover all the landmarks by applying physical force actions (equally discretized into eight directions) to move, where the entire 2-D working space is continuous and unbounded, but agents and landmarks are initialized in a $(-1,1)^2$ region as shown in Fig.~\ref{domains}c. 
At each timestep, the shared penalty is the negative sum of the landmark-nearest-agent-distances plus extra penalties for collisions. 
We use this domain to examine ROLA's performance when dealing with dense reward signals rather than the sparse case in the above two. 

\textbf{Antipodal Navigation.} We make this environment to be partially observable in the same way as described above while keeping anything else as the default in~\cite{CM3}, where each agent aims to cover an allocated target landmark (shown with the same color as the agent in Fig.~\ref{domains}d and receives an individual reward equaling to the negative distance to its target. 
The initial position of agents and their target landmarks are sampled in an antipodal configuration with a probability of 0.7, rather than always randomly initialized as in the above cooperative case. 
We want to check whether ROLA is still competitive under such a less collaborative scenario without global reward or not.      

\subsection{Algorithm Implementations}

We investigate ROLA's advantages against a broad set of state-of-the-art multi-agent actor-critic methods that are summarized in Table~\ref{baselines}. 

All methods use the same actor-net architecture consisting of one fully connected (FC) layer with Leaky-Relu as the activation function, one LSTM layer~\cite{LSTM} and one more FC layer as the output, where each layer has 64 units. 
Our critic-net implementation uses the same architecture across all methods, which contains two 64-unit FC layers with Leaky-Relu activation function and one FC output layer in the end, except, for IAC, an LSTM module replaces the middle layer due to its condition on local observations. For MAAC and VDN-based algorithms, we keep the critic's architecture as the original design. 
ROLA, IA2C, Central-V, and ECA handle explorations via a linear decaying $\epsilon$-soft policy and use $n$-step TD prediction, while other methods use the default exploration schemes and TD$(\lambda)$ in their published implementations.  

Different methods have different properties so that each method may require a particular set of hyper-parameters to achieve the best performance. 
Therefore, in training, we perform decent hyper-parameter tuning for each method over a wide range of hyper-parameters via exhaustive grid search and pick out the best performance of each method depending on its final converged value as the first priority and the learning speed as the second.   

\begin{figure*}[t!]
    \centering
    \captionsetup[subfigure]{labelformat=empty}
    \subcaptionbox{}
        [0.99\linewidth]{\includegraphics[height=0.42cm]{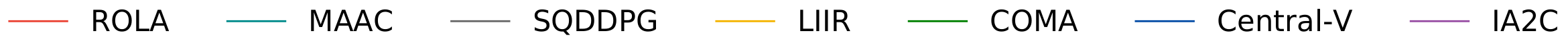}\vspace{-1mm}}
    \centering
    \subcaptionbox{\,\,\,\,\,\,\,(a) Capture Target\vspace{3mm}}
        [0.22\linewidth]{\includegraphics[height=3.6cm, width=4.2cm]{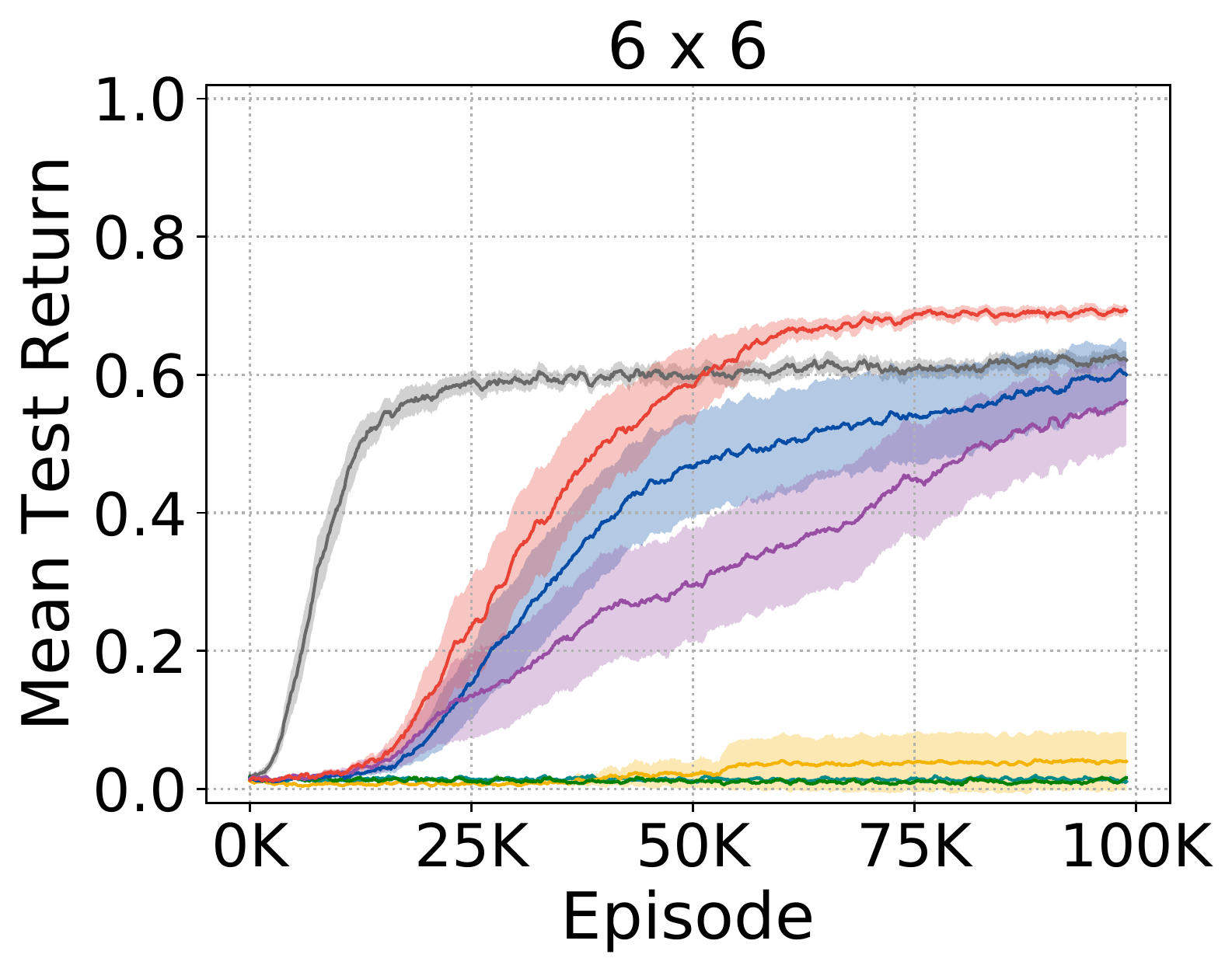}}
    ~
    \centering
    \subcaptionbox{\,\,\,\,\,\,\,(b) Capture Target}
        [0.22\linewidth]{\includegraphics[height=3.6cm, width=4.2cm]{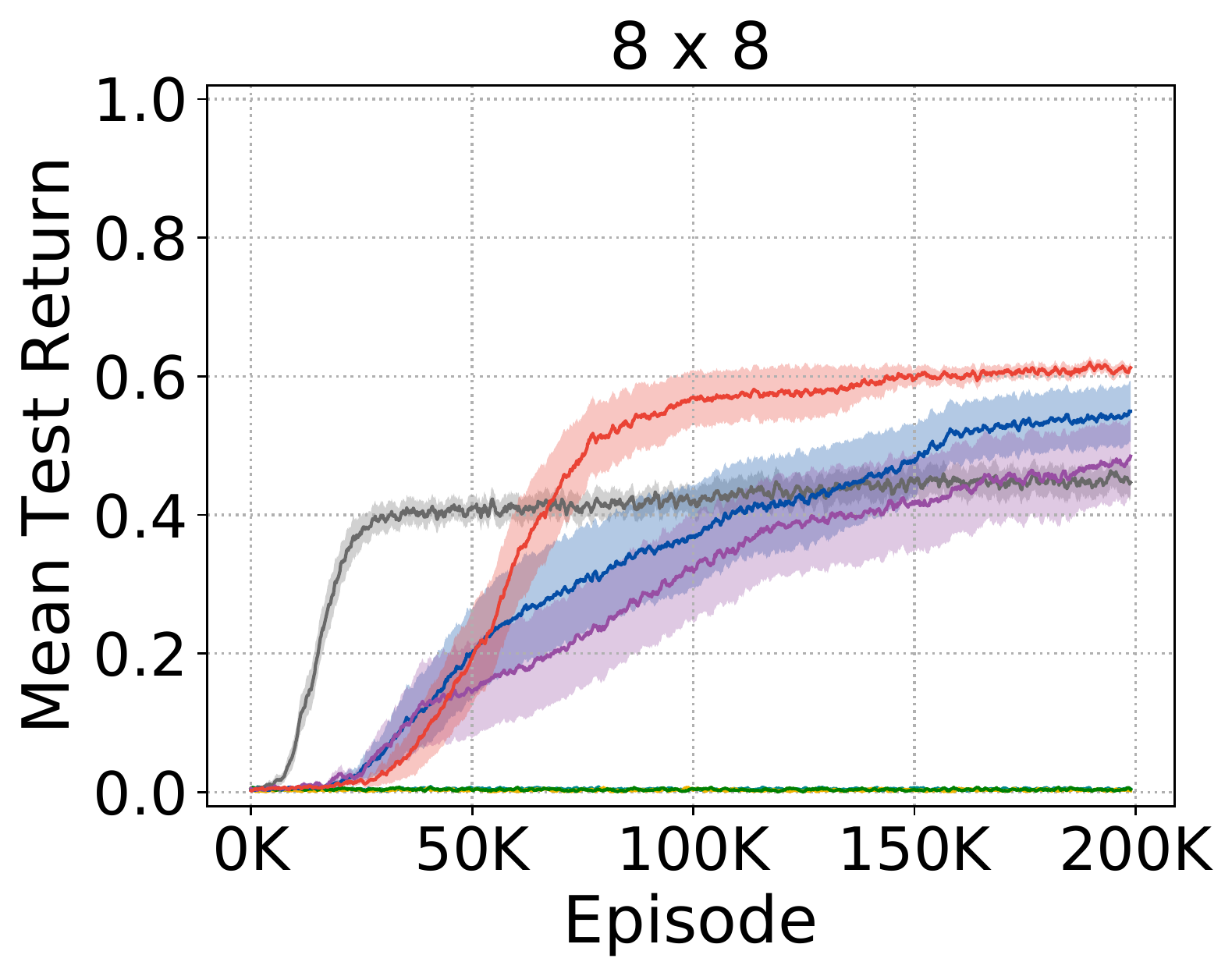}}
    ~
    \centering
    \subcaptionbox{\,\,\,\,\,\,\,(c) Box Pushing}
        [0.22\linewidth]{\includegraphics[height=3.6cm, width=4.2cm]{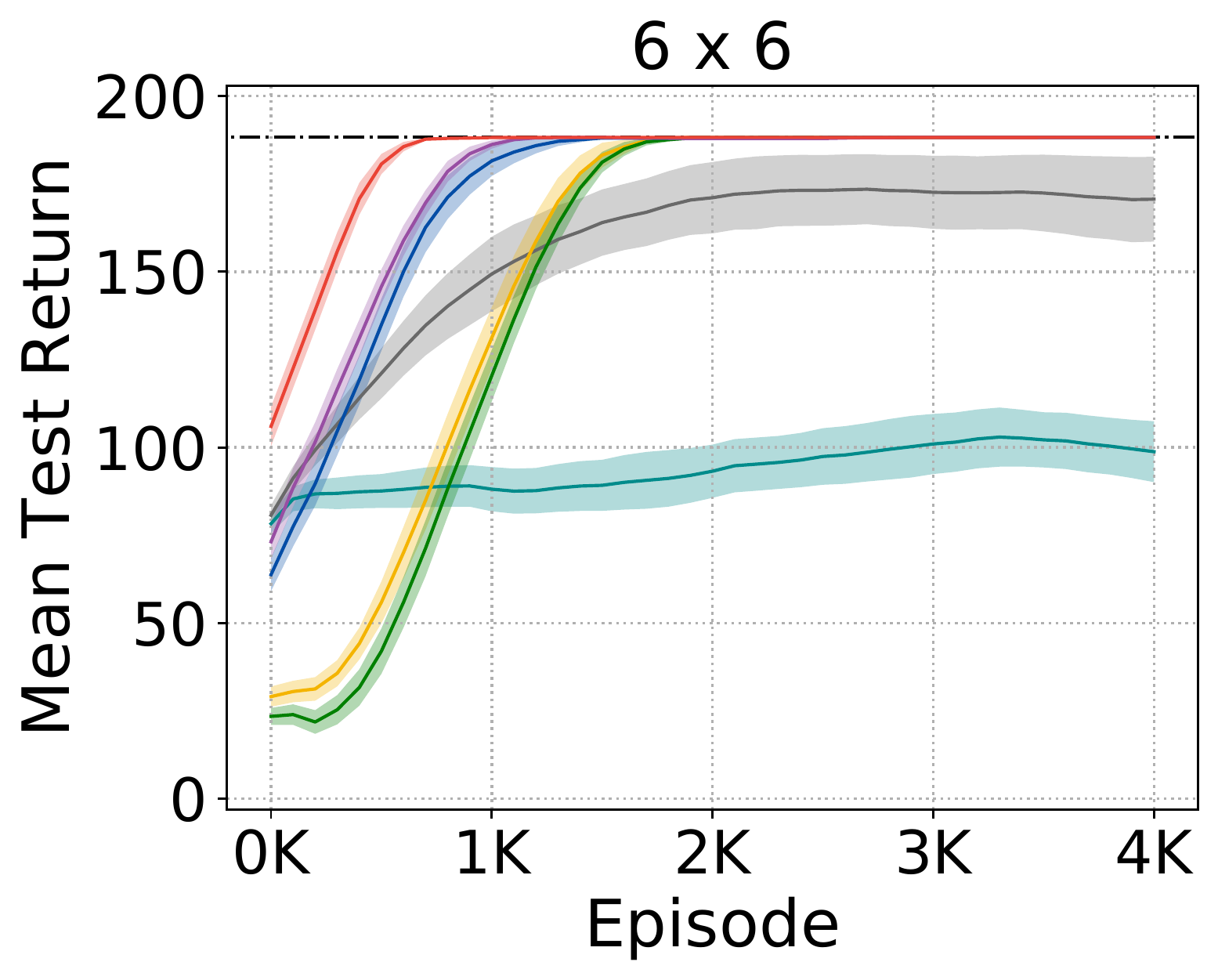}}
    ~
    \centering
    \subcaptionbox{\,\,\,\,\,\,\,(d) Box Pushing}
        [0.22\linewidth]{\includegraphics[height=3.6cm, width=4.2cm]{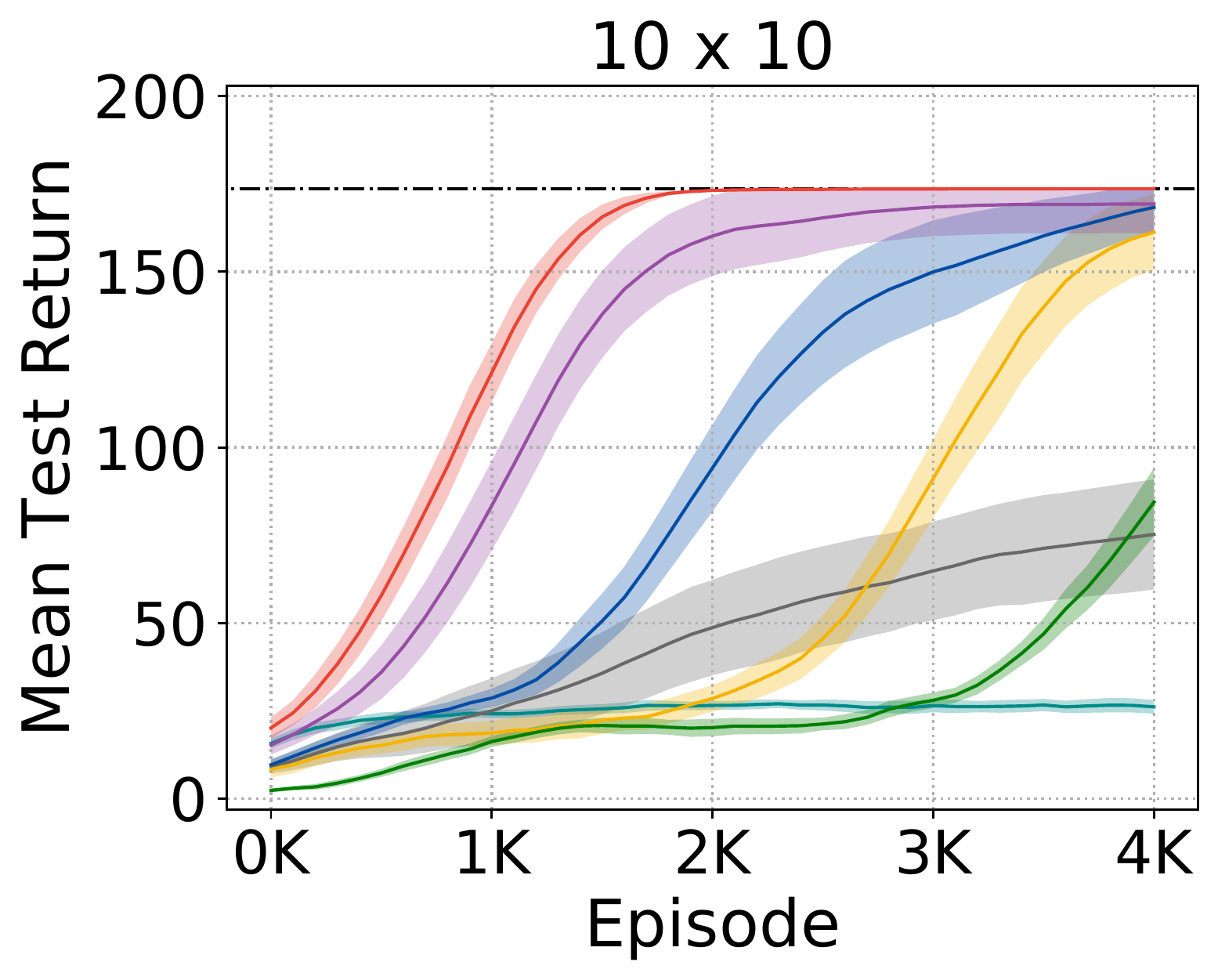}}
    \centering
    \subcaptionbox{\,\,\,\,\,\,\,\,(e) Cooperative Navigation\vspace{2mm}}
        [0.22\linewidth]{\includegraphics[height=3.6cm, width=4.3cm]{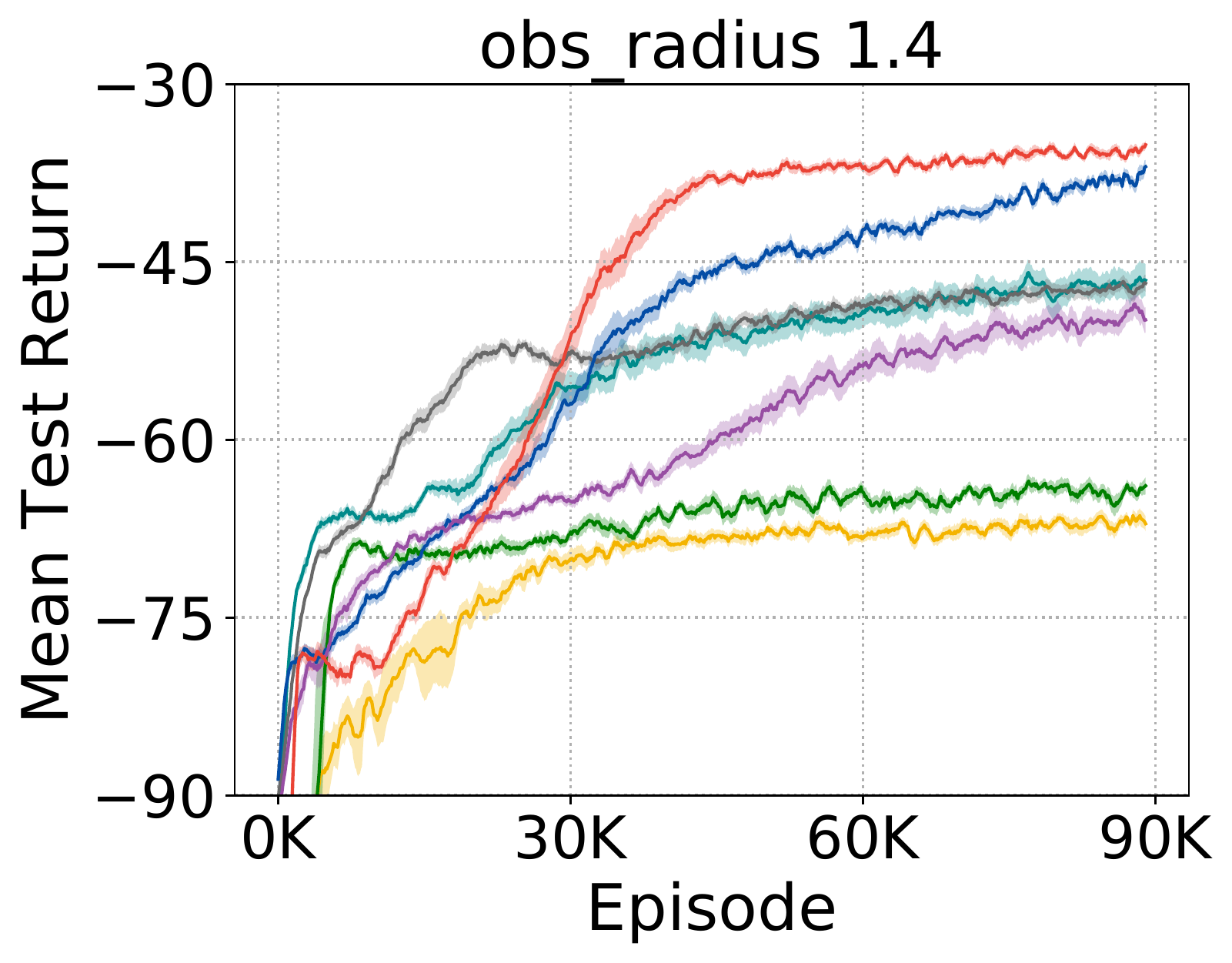}}
    ~
    \centering
    \subcaptionbox{\,\,\,\,\,\,\,\,(f) Cooperative Navigation}
        [0.22\linewidth]{\includegraphics[height=3.6cm, width=4.3cm]{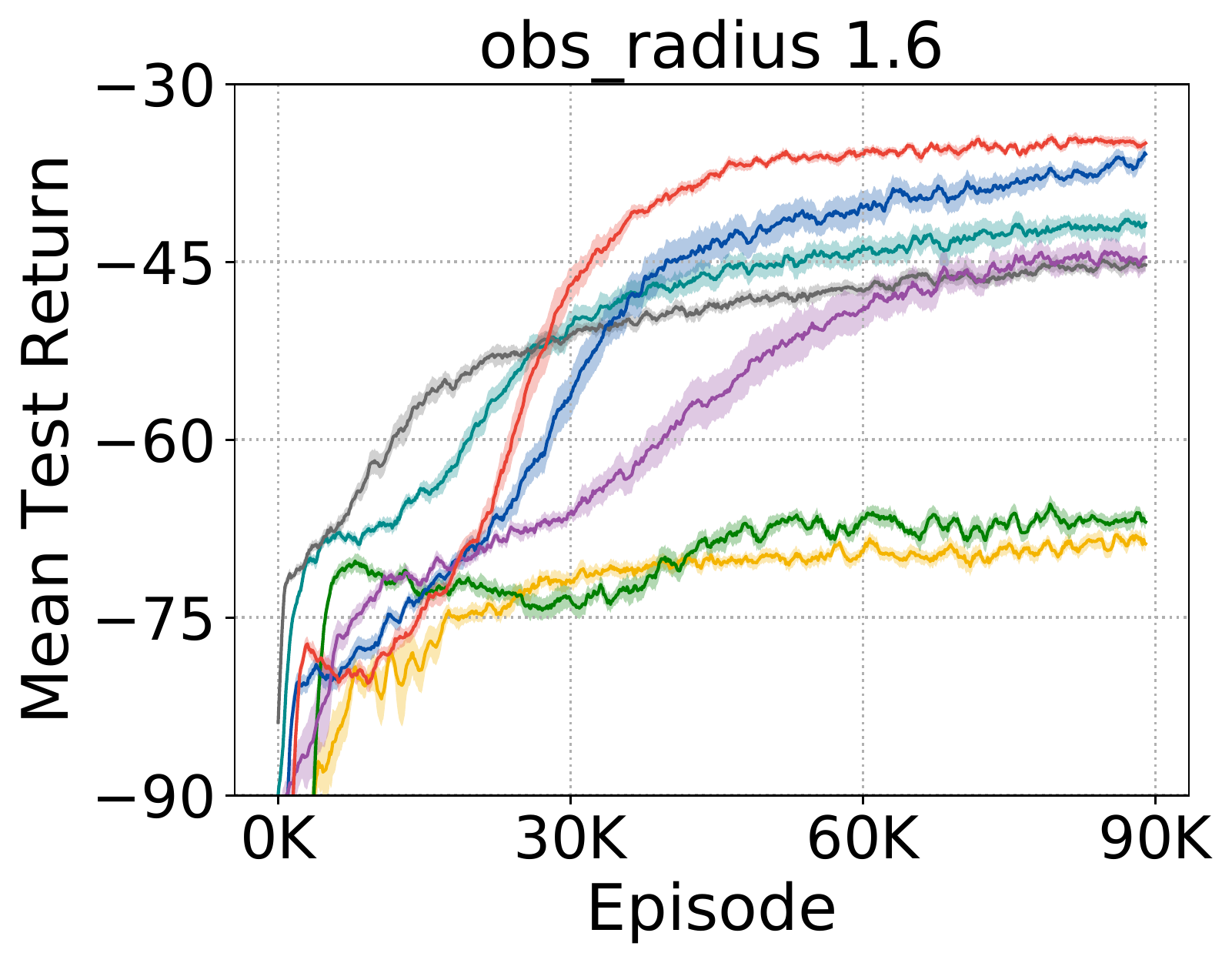}}
    ~
    \centering
    \subcaptionbox{\,\,\,\,\,\,\,\,\,\,\,\,\,(g) Antipodal Navigation}
        [0.22\linewidth]{\includegraphics[height=3.6cm, width=4.35cm]{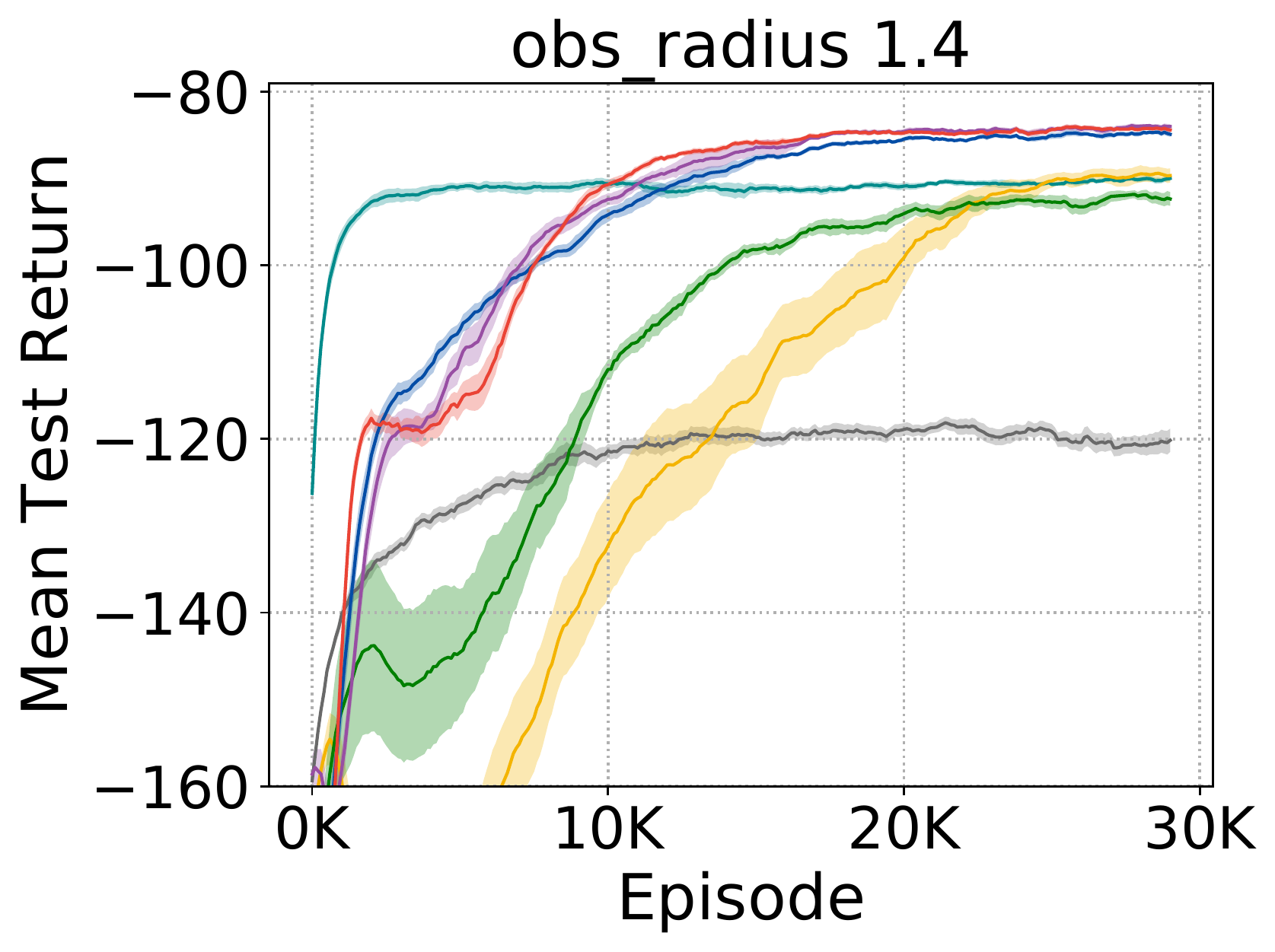}}
    ~
    \centering
    \subcaptionbox{\,\,\,\,\,\,\,\,\,\,\,\,\,(h) Antipodal Navigation}
        [0.23\linewidth]{\includegraphics[height=3.6cm, width=4.35cm]{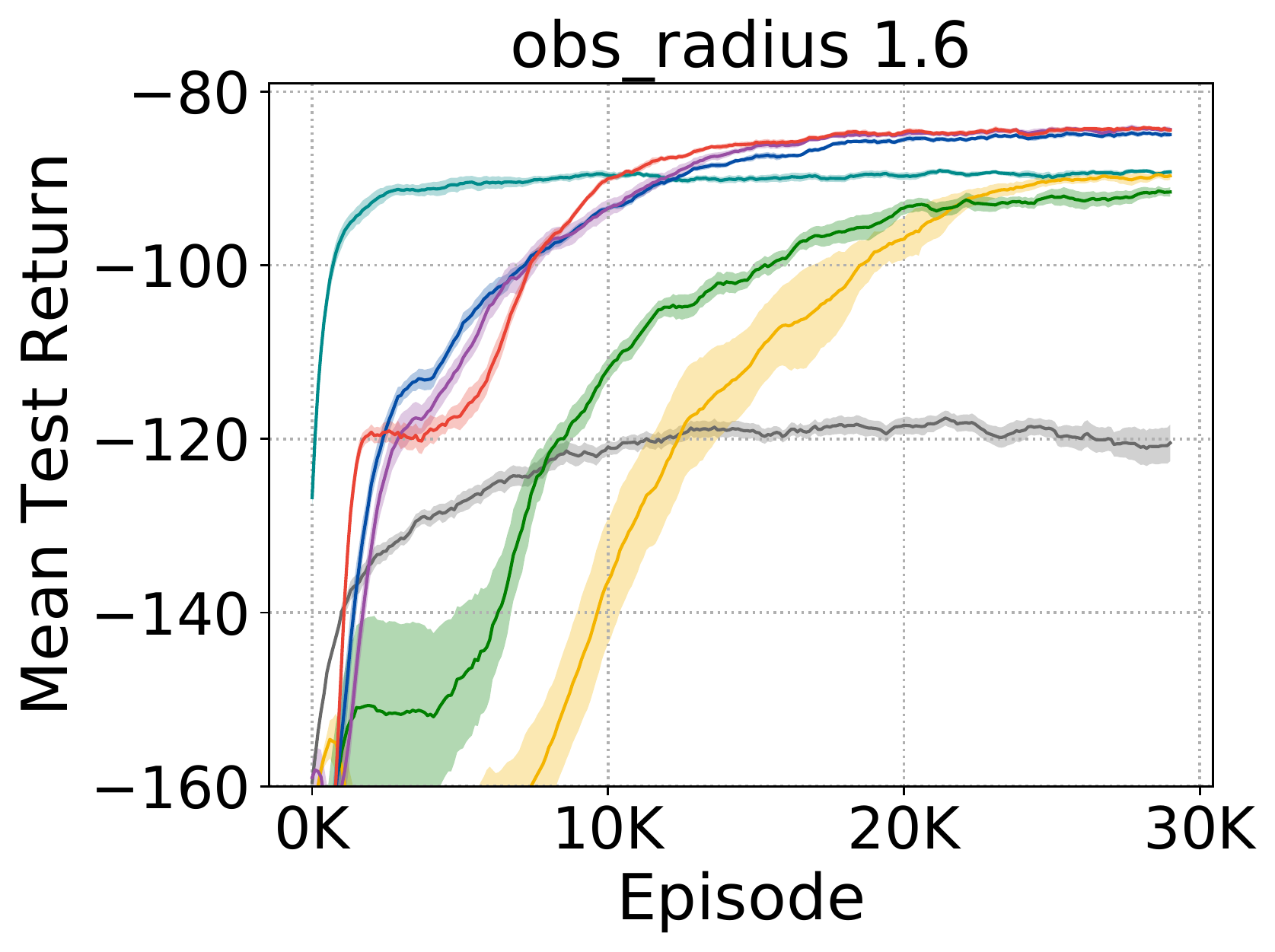}}
    \caption{Comparison against baseline algorithms in two scenarios of each domain.}
    \label{re1}
\end{figure*}

\subsection{Experimental Scheme}

We conduct experiments under each domain with modifications on task difficulty by either increasing the grid world size or decreasing the agent's observation range. 
For each method, we perform 20 independent training trials to solve all the tasks. 
Each training trial suspends every 100 episodes to evaluate the learned policies at the moment across another 10 episodes, which generates an averaged discounted return received by the multi-agent team over the evaluating episodes as one testing result. 
Finally, we calculate the mean testing performance of each method over the 20 trials and plot the curves (smoothed by a window size 10) with a $95\%$ confidence interval.   

\section{Results and Discussions}

In this section, we report the performance of ROLA against all the baselines on two environmental configurations under each domain. 
The comparisons between ROLA and the baselines are organized as the following pattern:

\emph{\textbf{Comparison 1}} (Fig.~\ref{re1}). We first compare ROLA with IA2C, Central-V, COMA, LIIR, SQDDPG, and MAAC, in order to: 
(a) examine the effectiveness of ROLA on credit assignment and variance reduction; 
(b) check if ROLA has strong robustness such that it can consistently outstand the baselines across different domains.     

\begin{figure*}[t!]
    \centering
    \captionsetup[subfigure]{labelformat=empty}
    \subcaptionbox{}
        [0.98\linewidth]{\includegraphics[height=0.43cm]{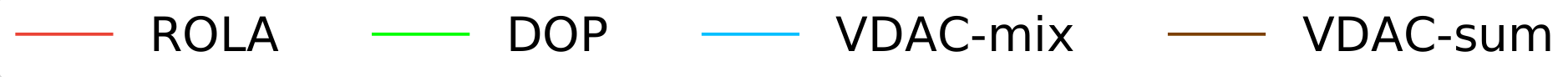}\vspace{-1mm}}
    \centering
    \subcaptionbox{\,\,\,\,\,\,\,(a) Capture Target\vspace{3mm}}
        [0.22\linewidth]{\includegraphics[height=3.6cm, width=4.2cm]{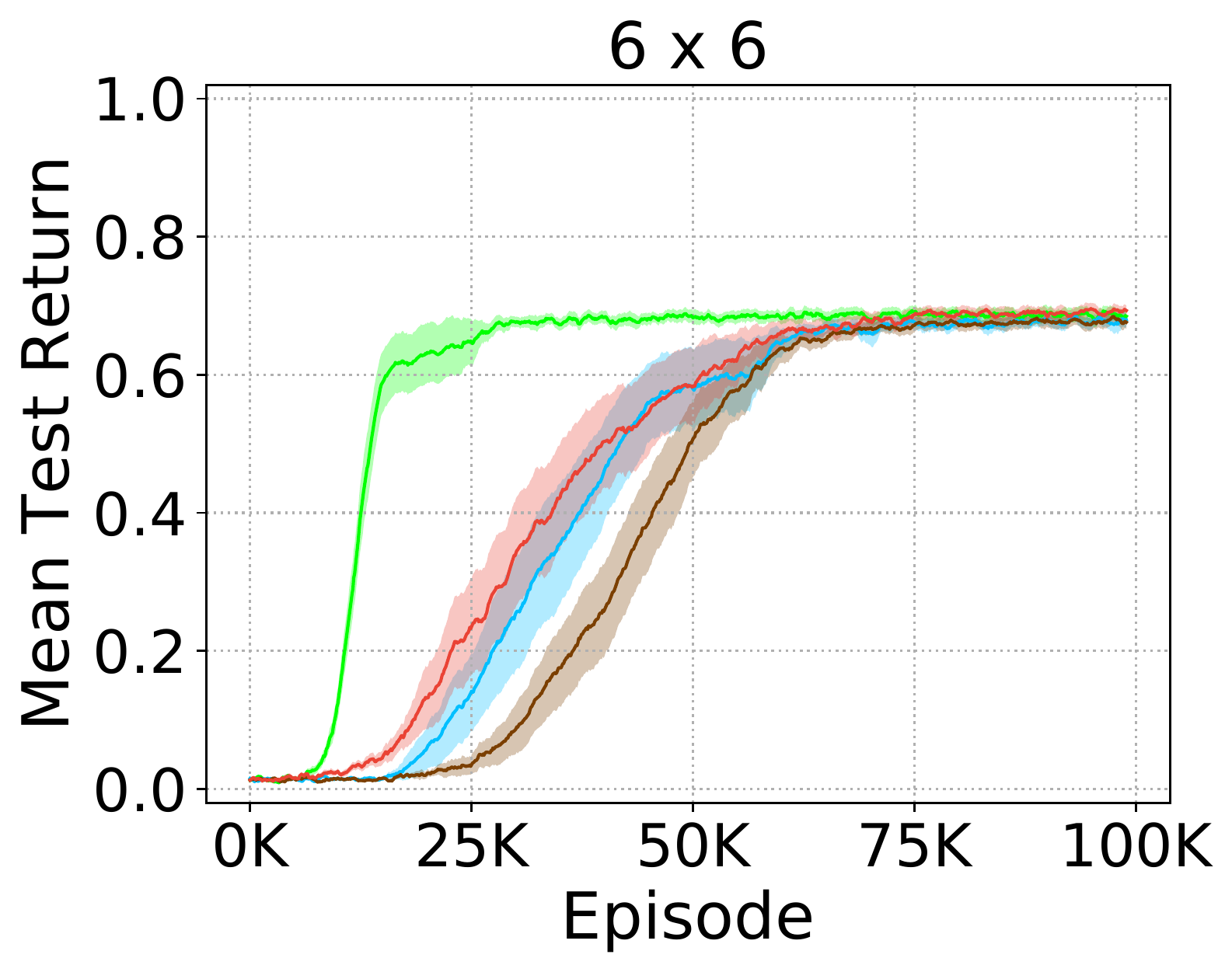}}
    ~
    \centering
    \subcaptionbox{\,\,\,\,\,\,\,(b) Capture Target}
        [0.22\linewidth]{\includegraphics[height=3.6cm, width=4.2cm]{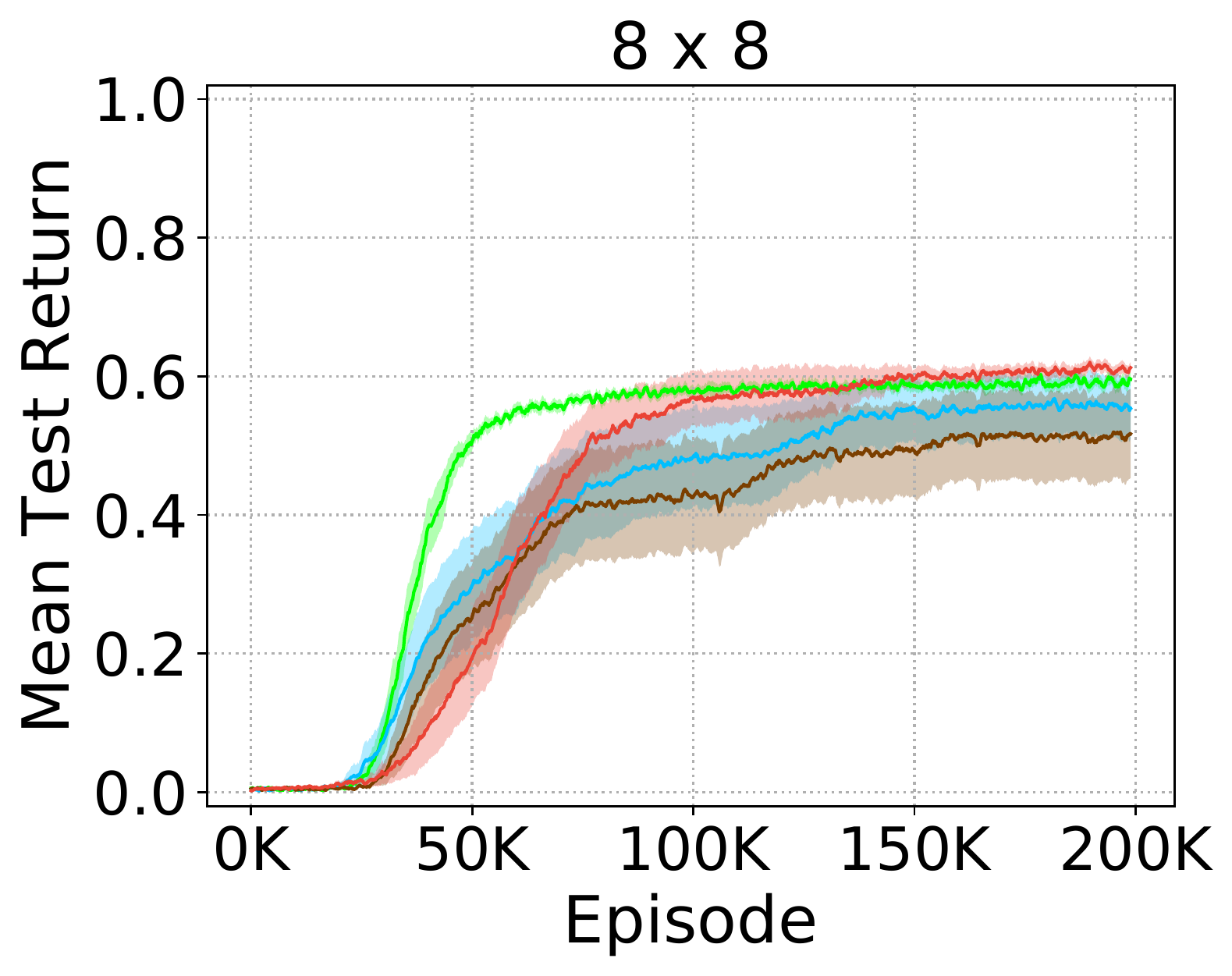}}
    ~
    \centering
    \subcaptionbox{\,\,\,\,\,\,\,(c) Box Pushing}
        [0.22\linewidth]{\includegraphics[height=3.6cm, width=4.2cm]{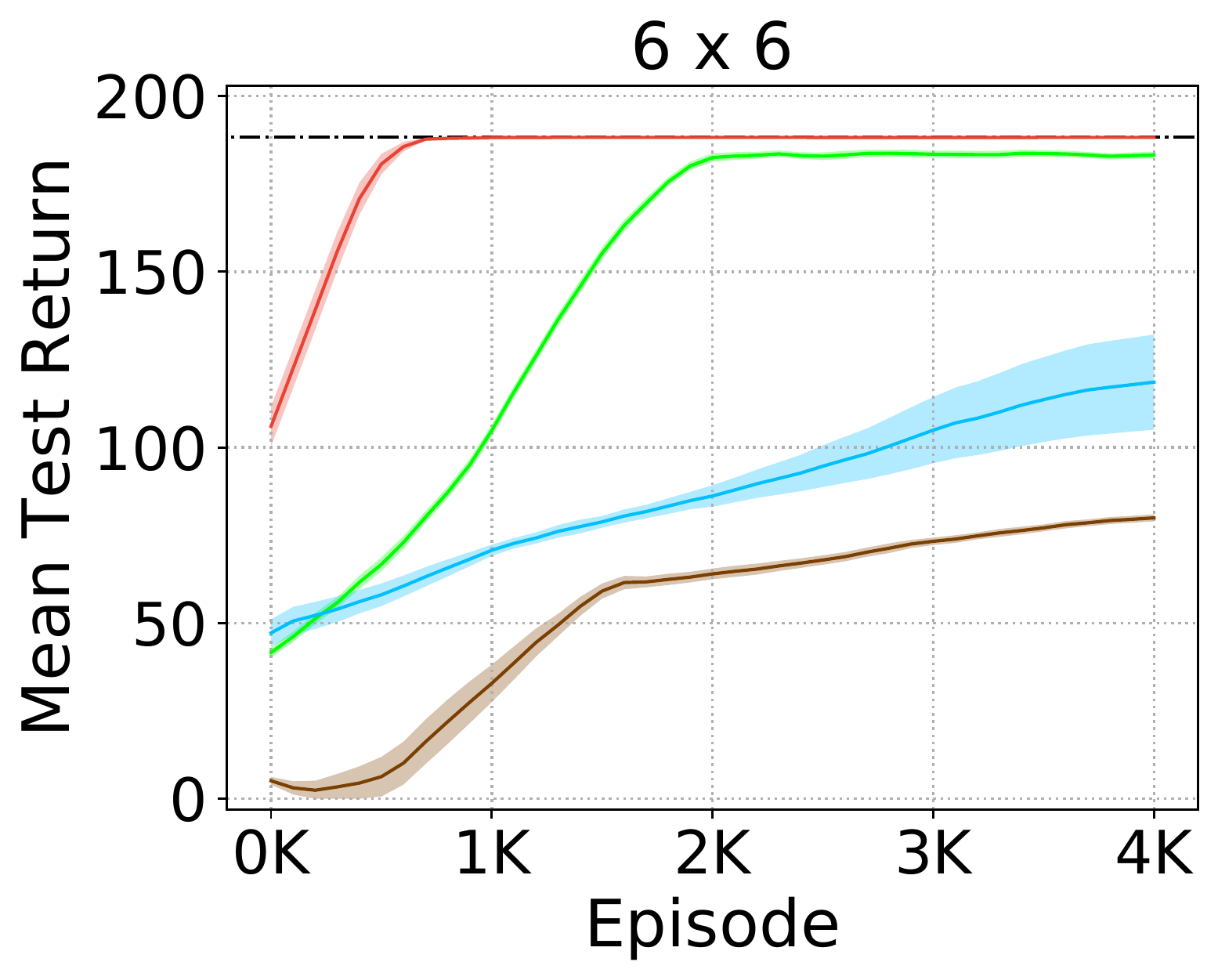}}
    ~
    \centering
    \subcaptionbox{\,\,\,\,\,\,\,(d) Box Pushing}
        [0.22\linewidth]{\includegraphics[height=3.6cm, width=4.2cm]{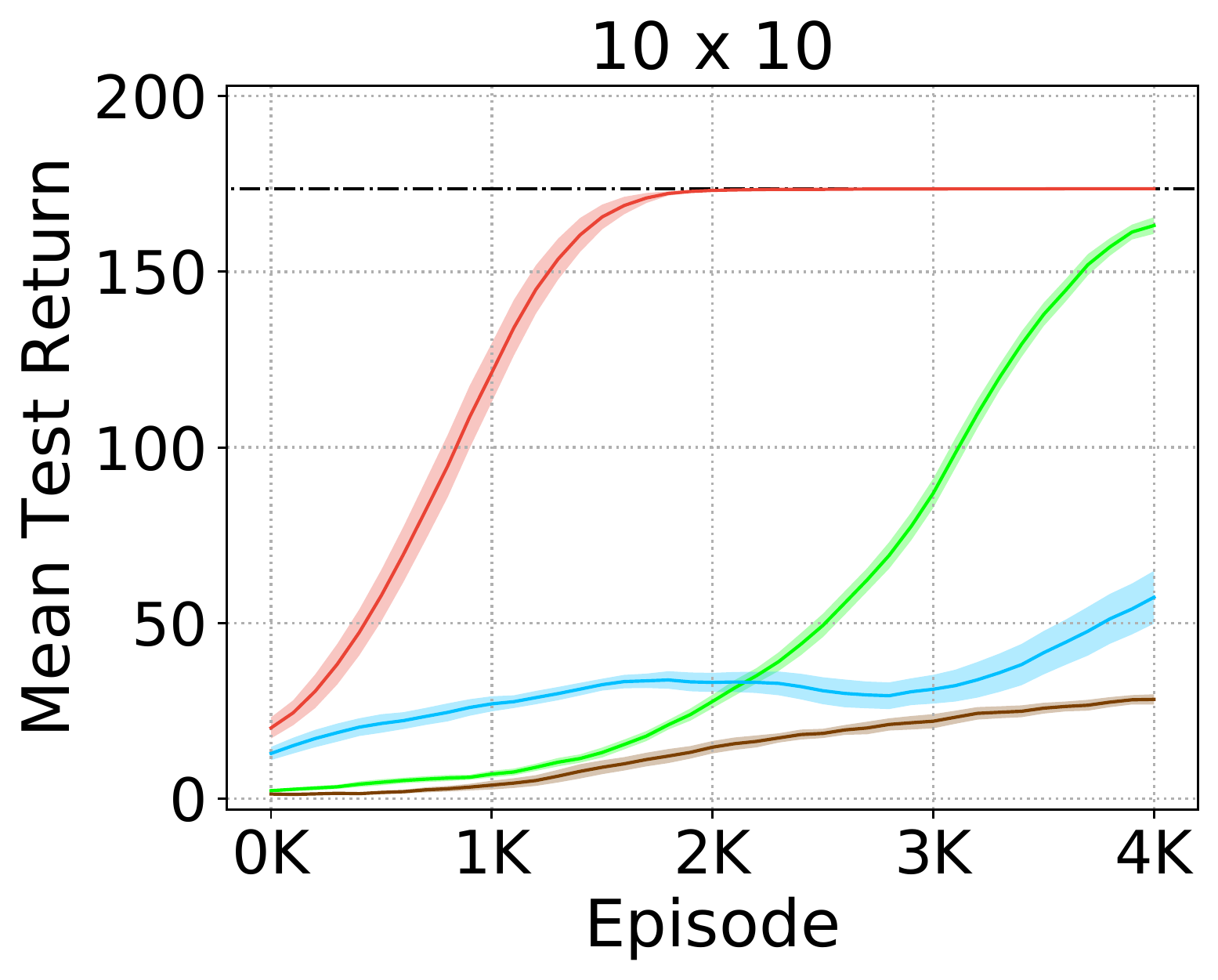}}
    \centering
    \subcaptionbox{\,\,\,\,\,\,\,\,(e) Cooperative Navigation\vspace{2mm}}
        [0.22\linewidth]{\includegraphics[height=3.6cm, width=4.3cm]{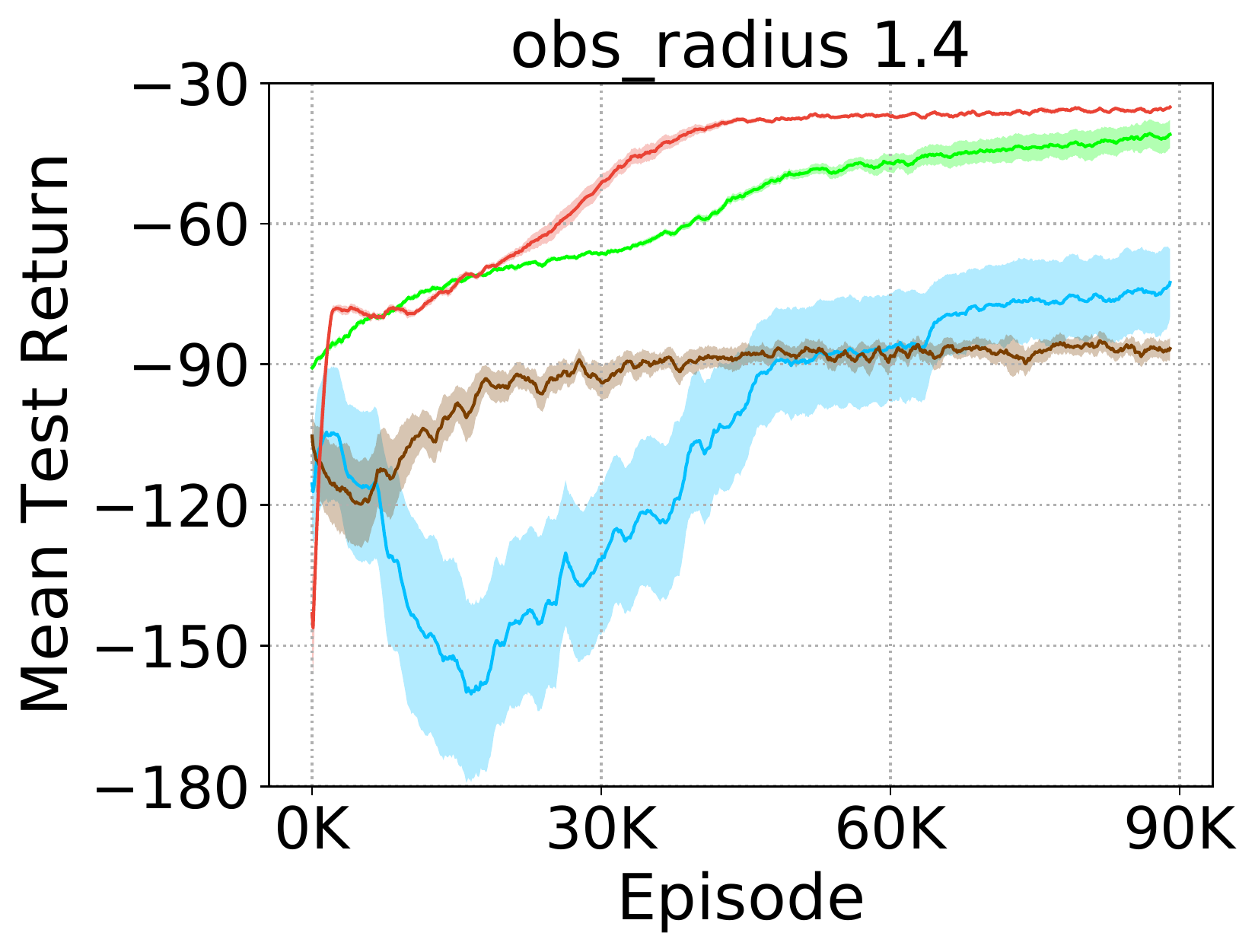}}
    ~
    \centering
    \subcaptionbox{\,\,\,\,\,\,\,\,(f) Cooperative Navigation}
        [0.22\linewidth]{\includegraphics[height=3.6cm, width=4.3cm]{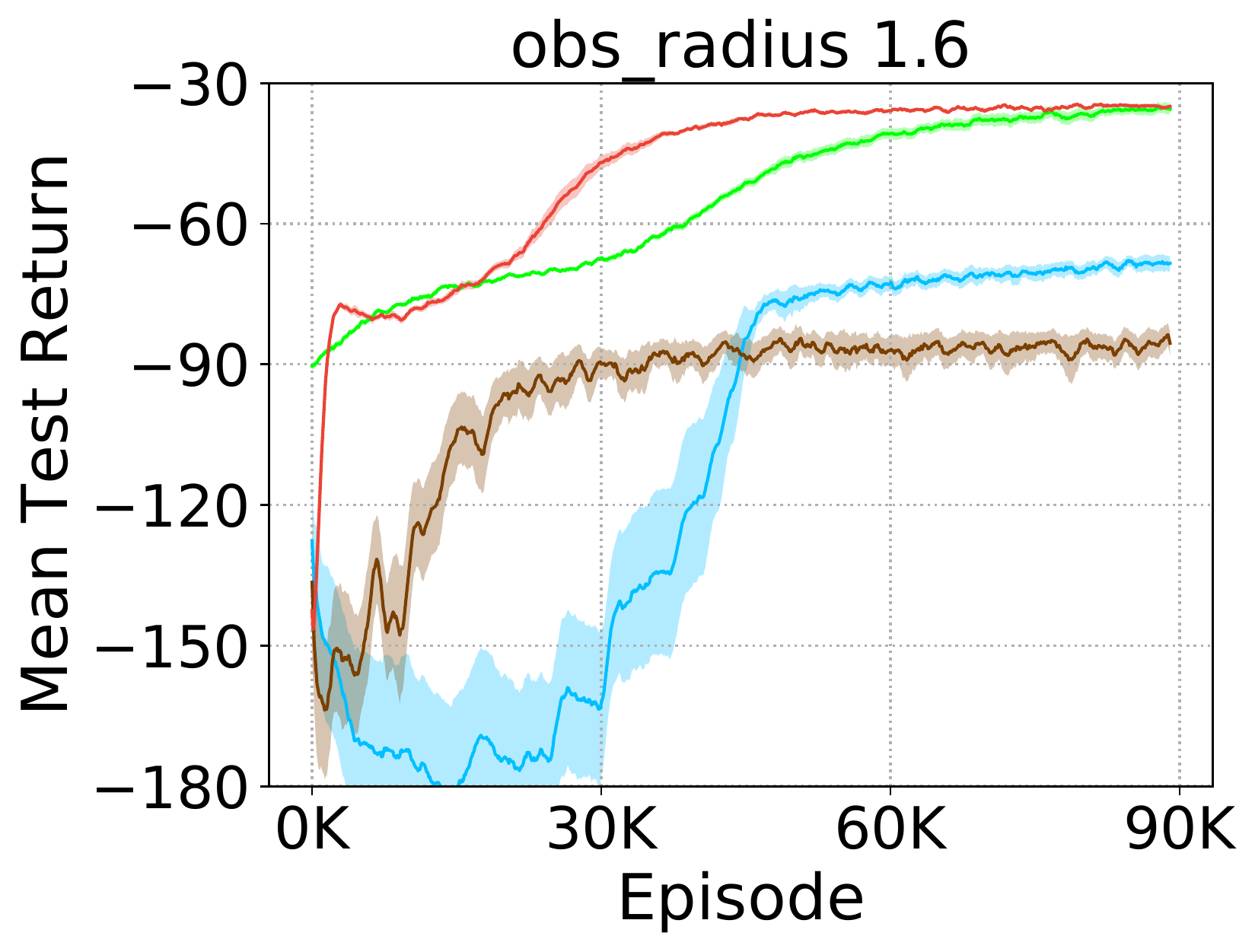}}
    ~
    \centering
    \subcaptionbox{\,\,\,\,\,\,\,\,\,\,\,\,\,(g) Antipodal Navigation}
        [0.22\linewidth]{\includegraphics[height=3.6cm, width=4.35cm]{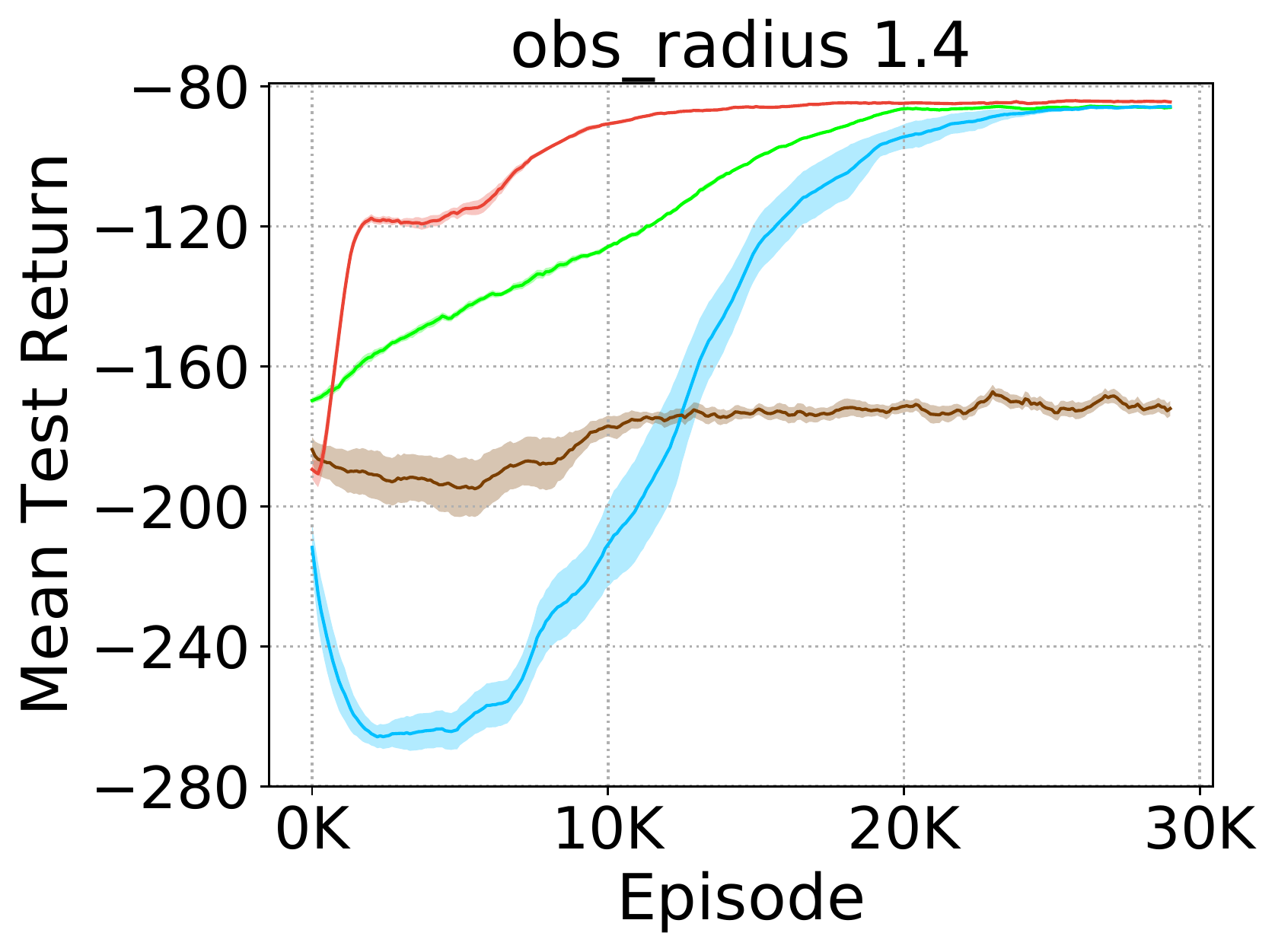}}
    ~
    \centering
    \subcaptionbox{\,\,\,\,\,\,\,\,\,\,\,\,\,(h) Antipodal Navigation}
        [0.23\linewidth]{\includegraphics[height=3.6cm, width=4.35cm]{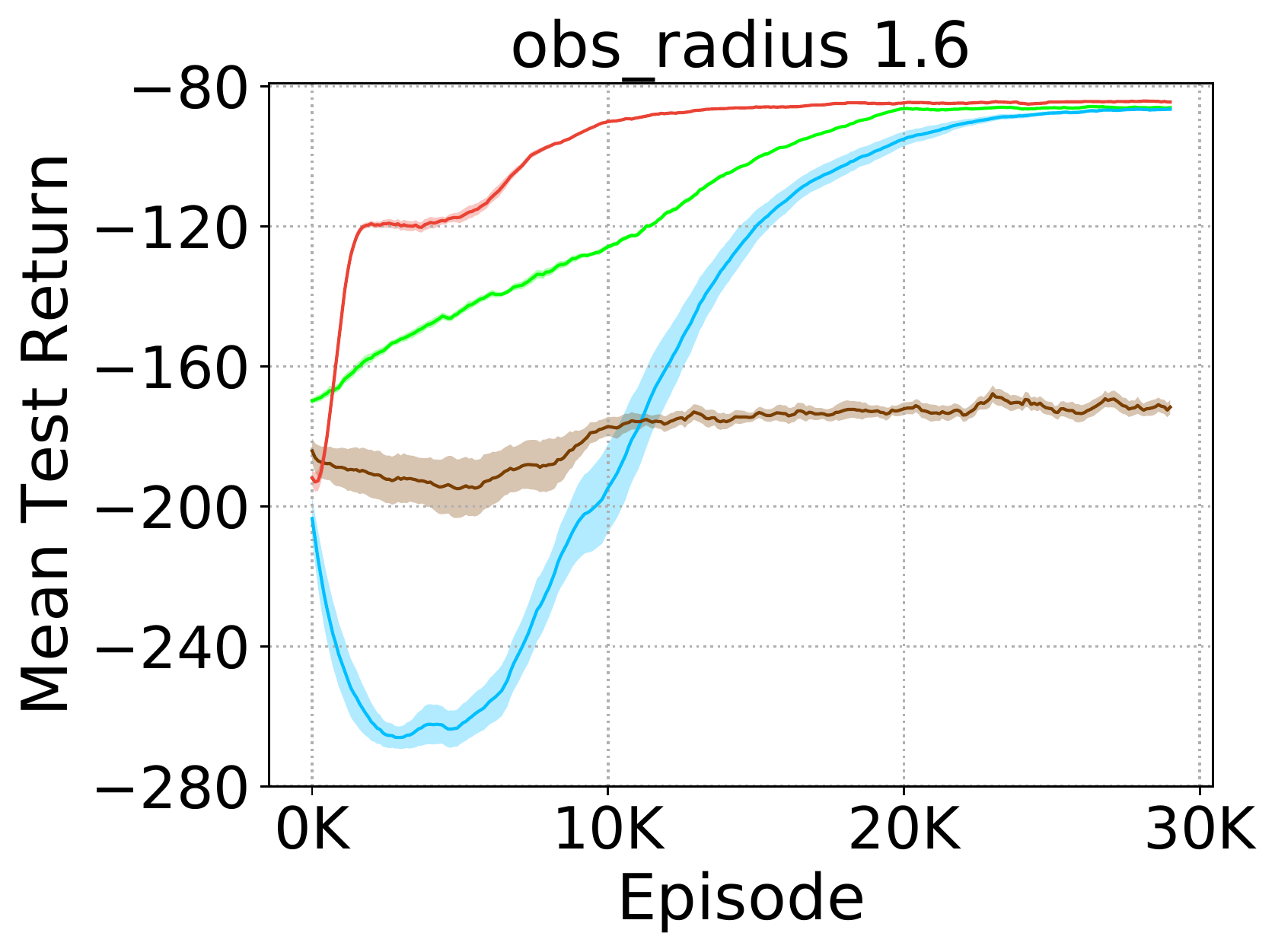}}
    \caption{Comparison against VDN-based algorithms in two scenarios of each domain.}
    \label{re2}
\end{figure*}

\emph{\textbf{Comparison 2}} (Fig.~\ref{re2}). Here, we compare ROLA with DOP, VDAC-mix, and VDAC-sum, intending to study ROLA's advance on learning local critics against value decomposition-based manners.

\begin{figure*}[t!]
    \centering
    \captionsetup[subfigure]{labelformat=empty}
    \subcaptionbox{}
        [0.98\linewidth]{\includegraphics[height=0.43cm]{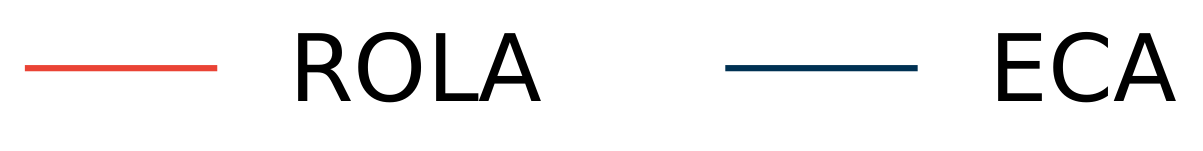}\vspace{-1mm}}
    \centering
    \subcaptionbox{\,\,\,\,\,\,\,(a) Capture Target\vspace{3mm}}
        [0.22\linewidth]{\includegraphics[height=3.6cm, width=4.2cm]{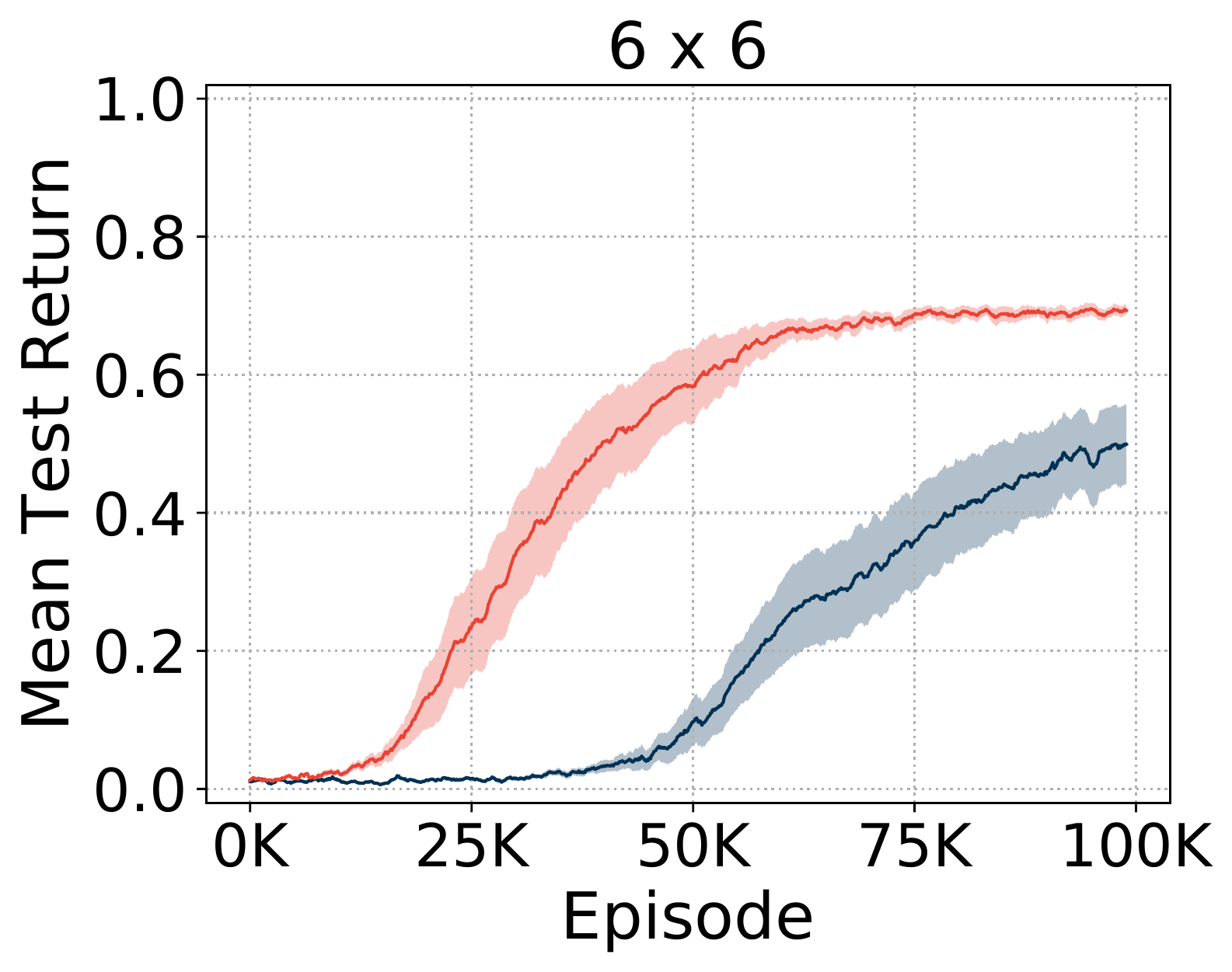}}
    ~
    \centering
    \subcaptionbox{\,\,\,\,\,\,\,(b) Capture Target}
        [0.22\linewidth]{\includegraphics[height=3.6cm, width=4.2cm]{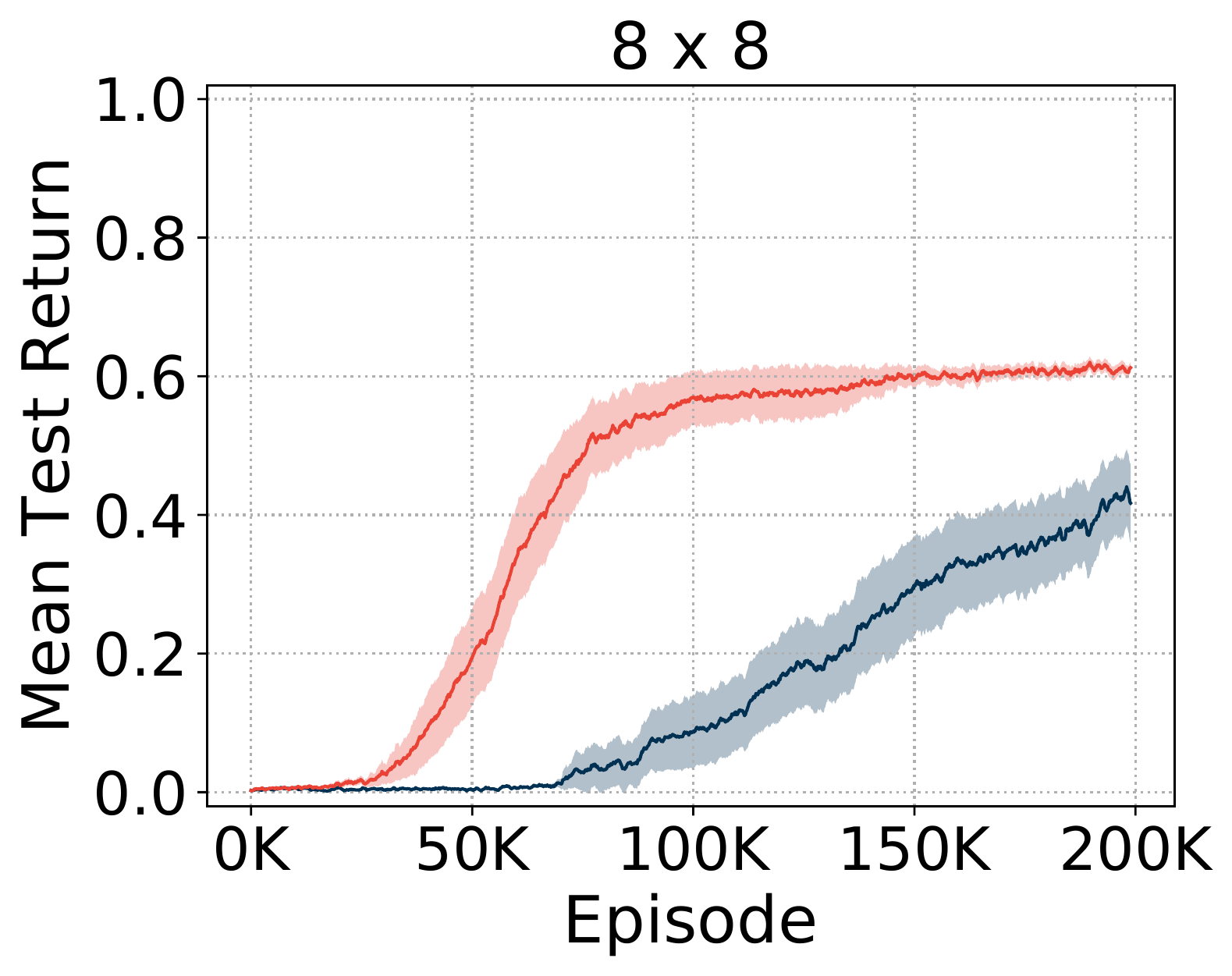}}
    ~
    \centering
    \subcaptionbox{\,\,\,\,\,\,\,(c) Box Pushing}
        [0.22\linewidth]{\includegraphics[height=3.6cm, width=4.2cm]{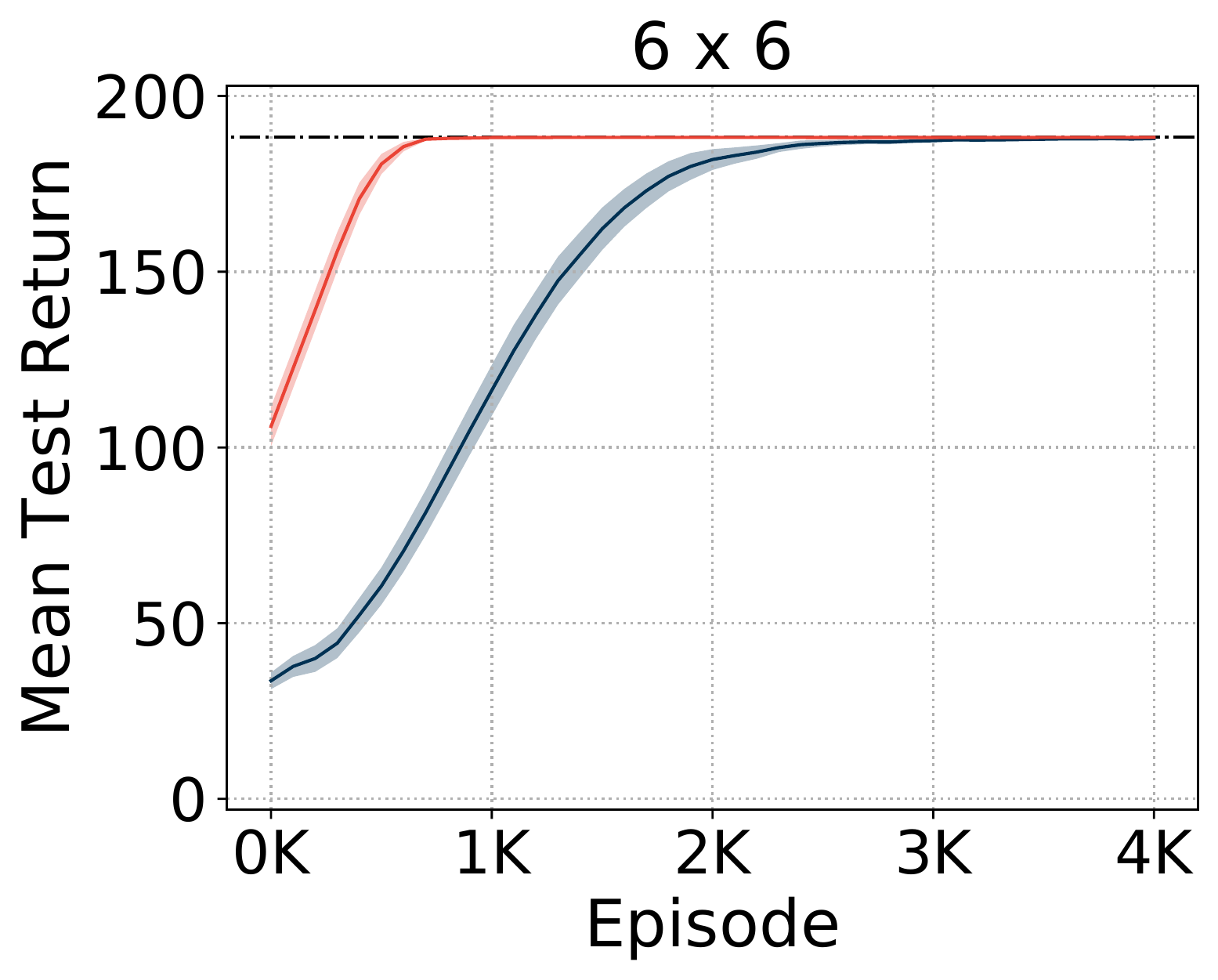}}
    ~
    \centering
    \subcaptionbox{\,\,\,\,\,\,\,(d) Box Pushing}
        [0.22\linewidth]{\includegraphics[height=3.6cm, width=4.2cm]{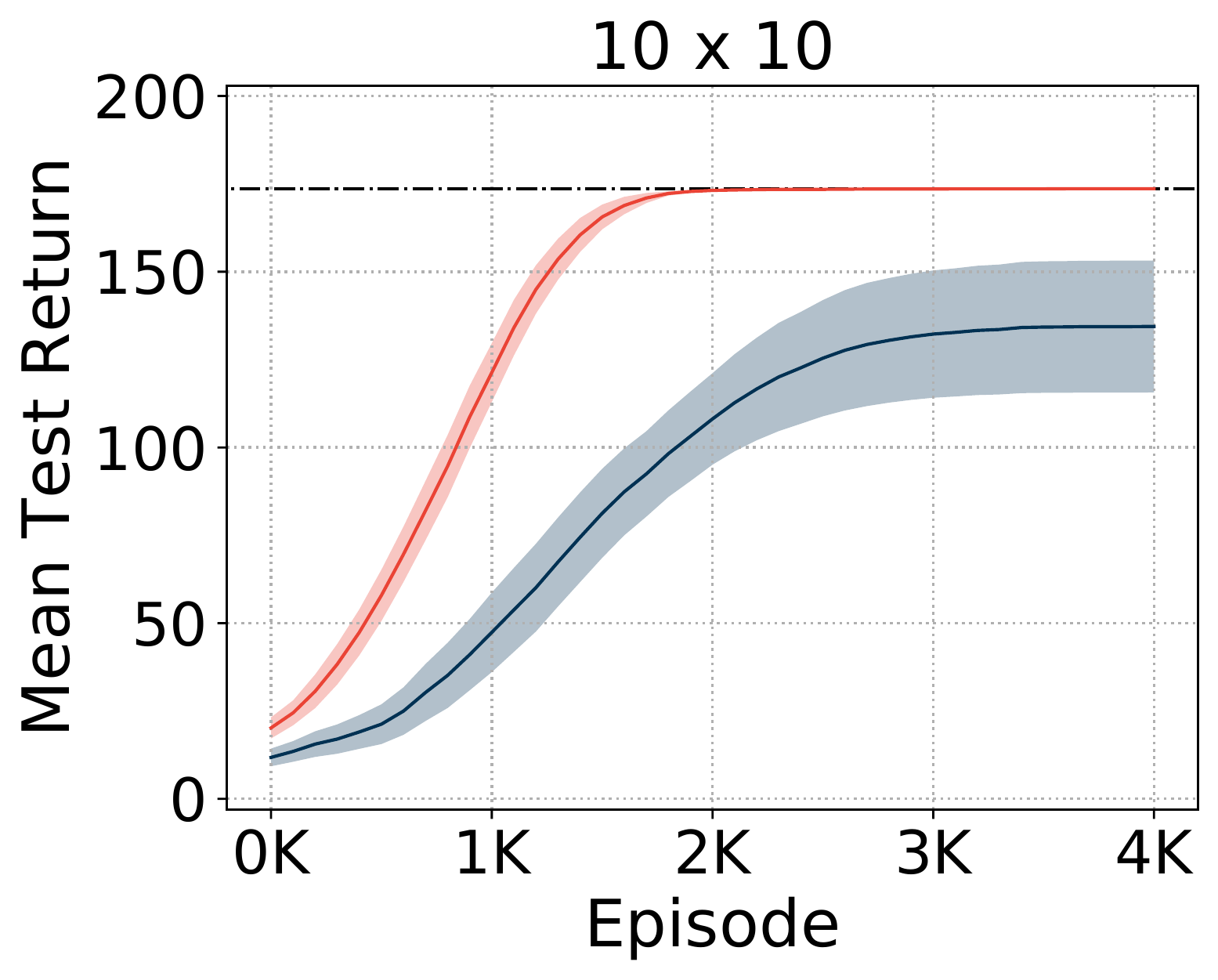}}
    \centering
    \subcaptionbox{\,\,\,\,\,\,\,\,(e) Cooperative Navigation\vspace{2mm}}
        [0.22\linewidth]{\includegraphics[height=3.6cm, width=4.3cm]{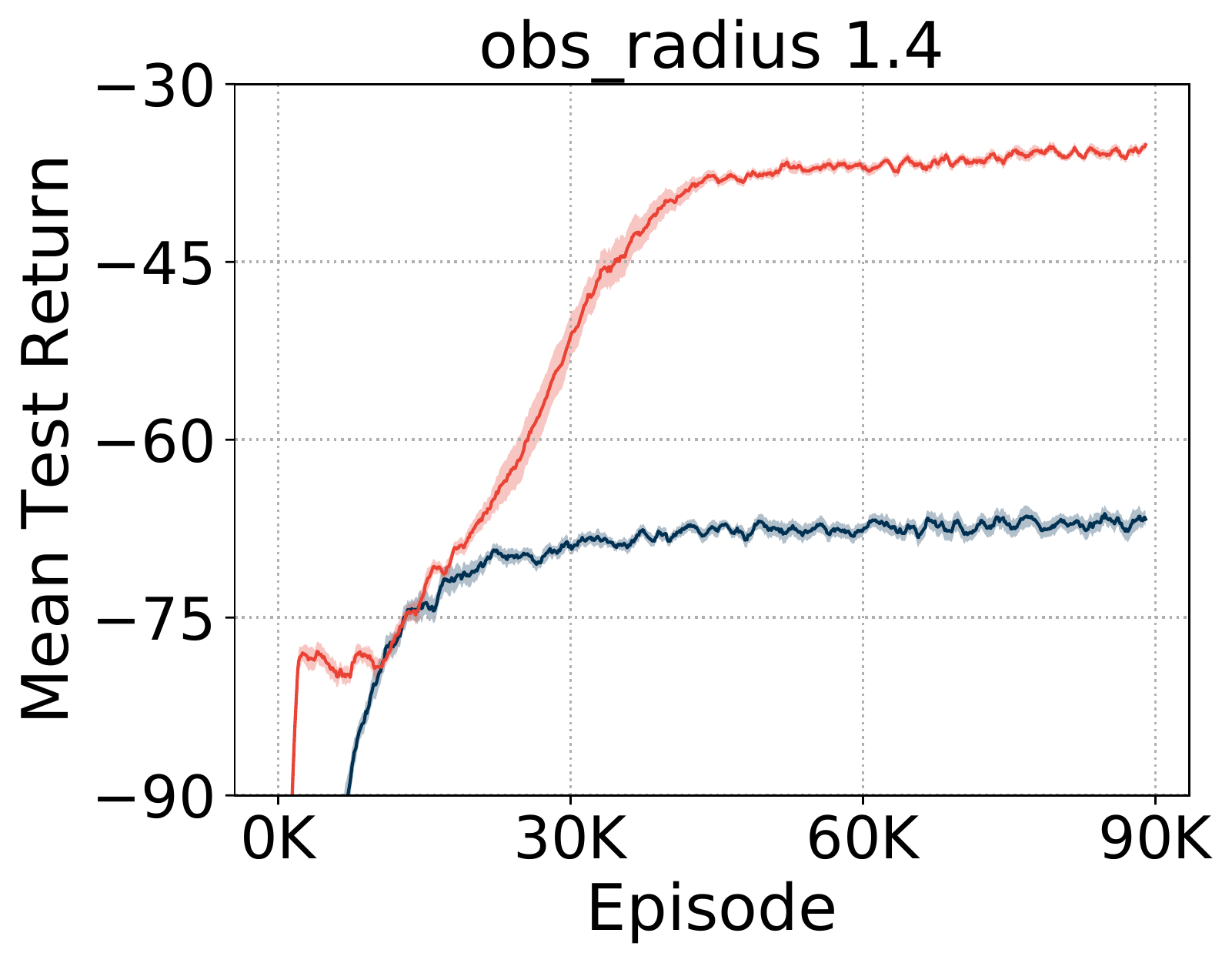}}
    ~
    \centering
    \subcaptionbox{\,\,\,\,\,\,\,\,(f) Cooperative Navigation}
        [0.22\linewidth]{\includegraphics[height=3.6cm, width=4.3cm]{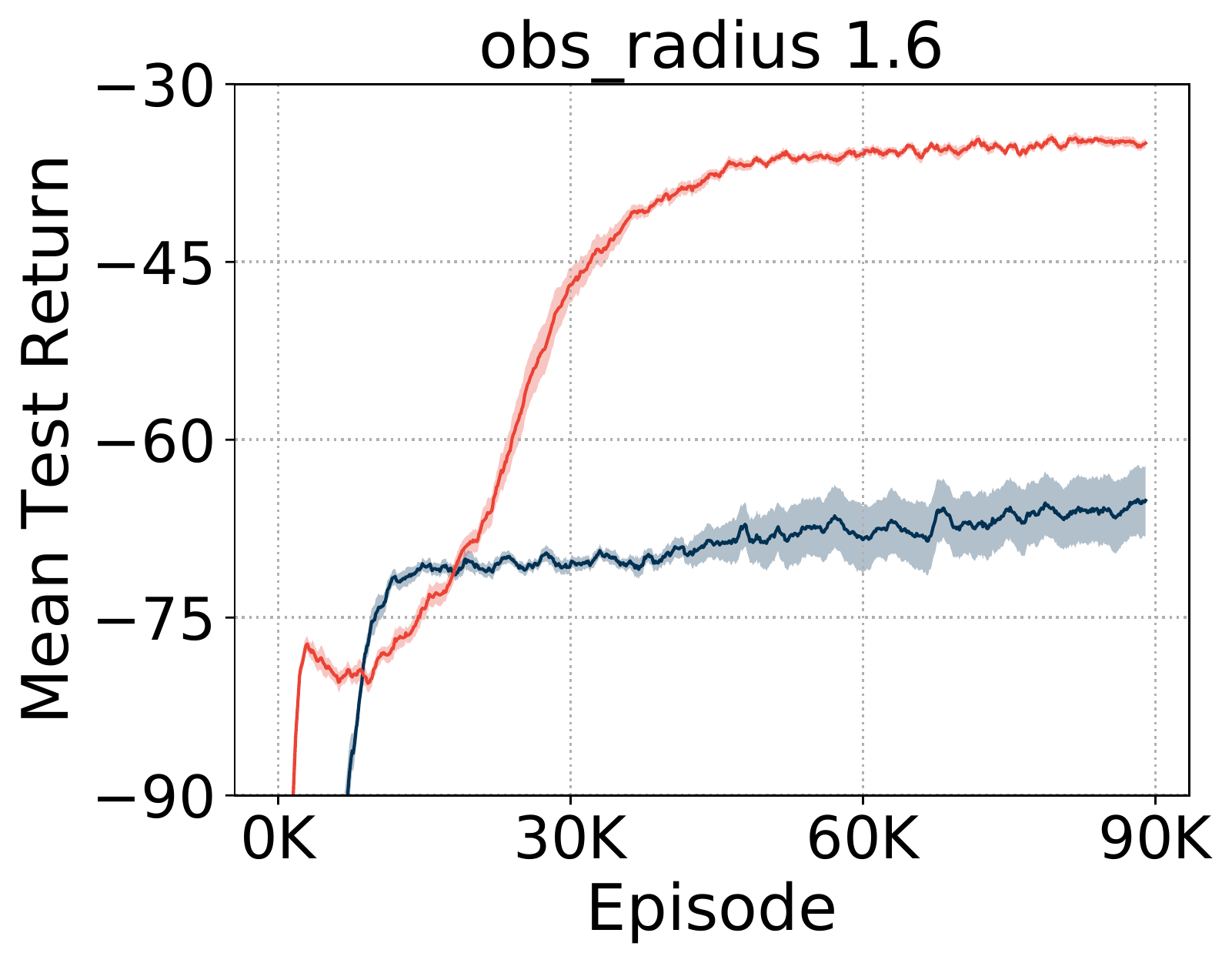}}
    ~
    \centering
    \subcaptionbox{\,\,\,\,\,\,\,\,\,\,\,\,\,(g) Antipodal Navigation}
        [0.22\linewidth]{\includegraphics[height=3.6cm, width=4.35cm]{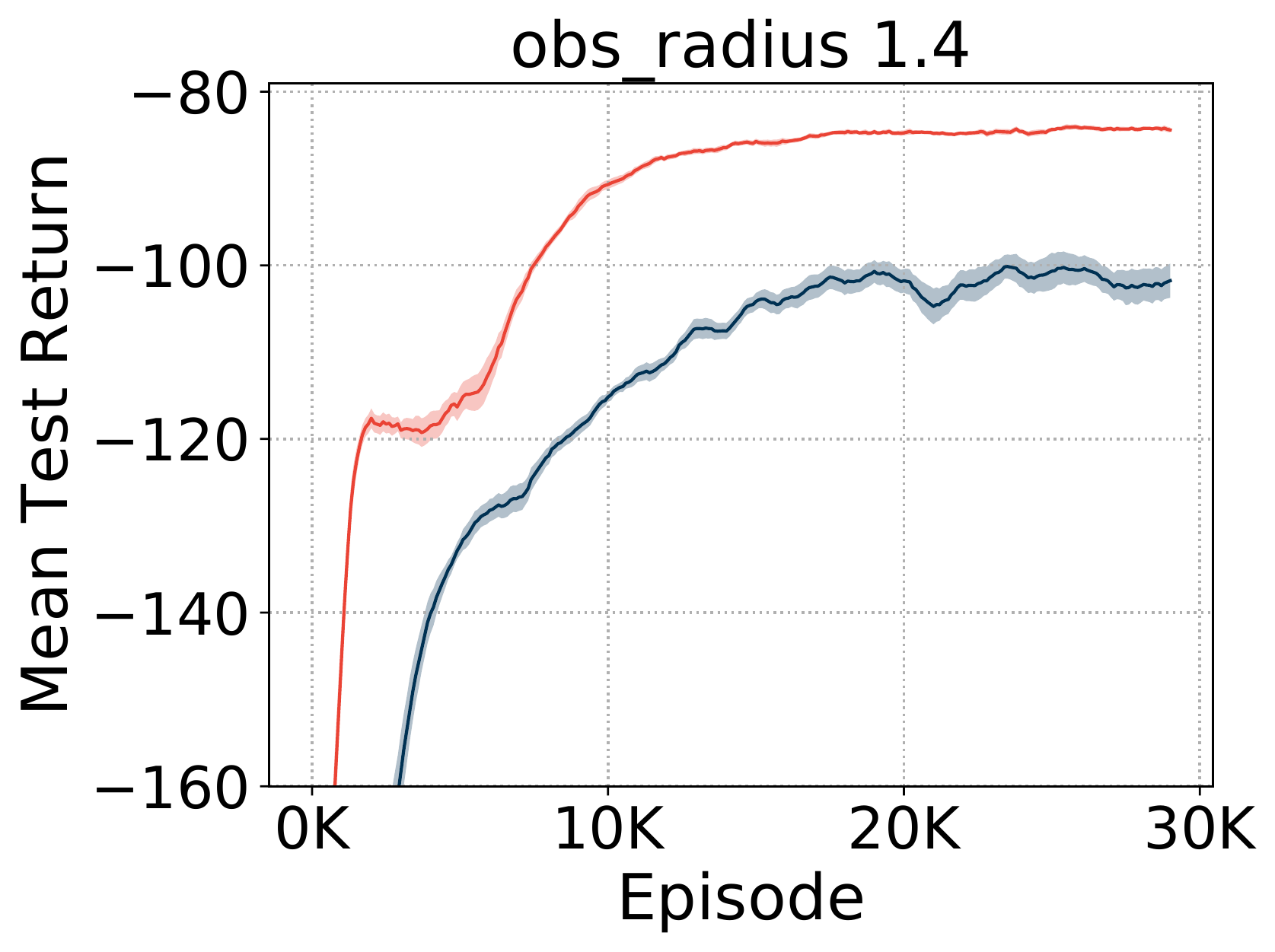}}
    ~
    \centering
    \subcaptionbox{\,\,\,\,\,\,\,\,\,\,\,\,\,(h) Antipodal Navigation}
        [0.23\linewidth]{\includegraphics[height=3.6cm, width=4.35cm]{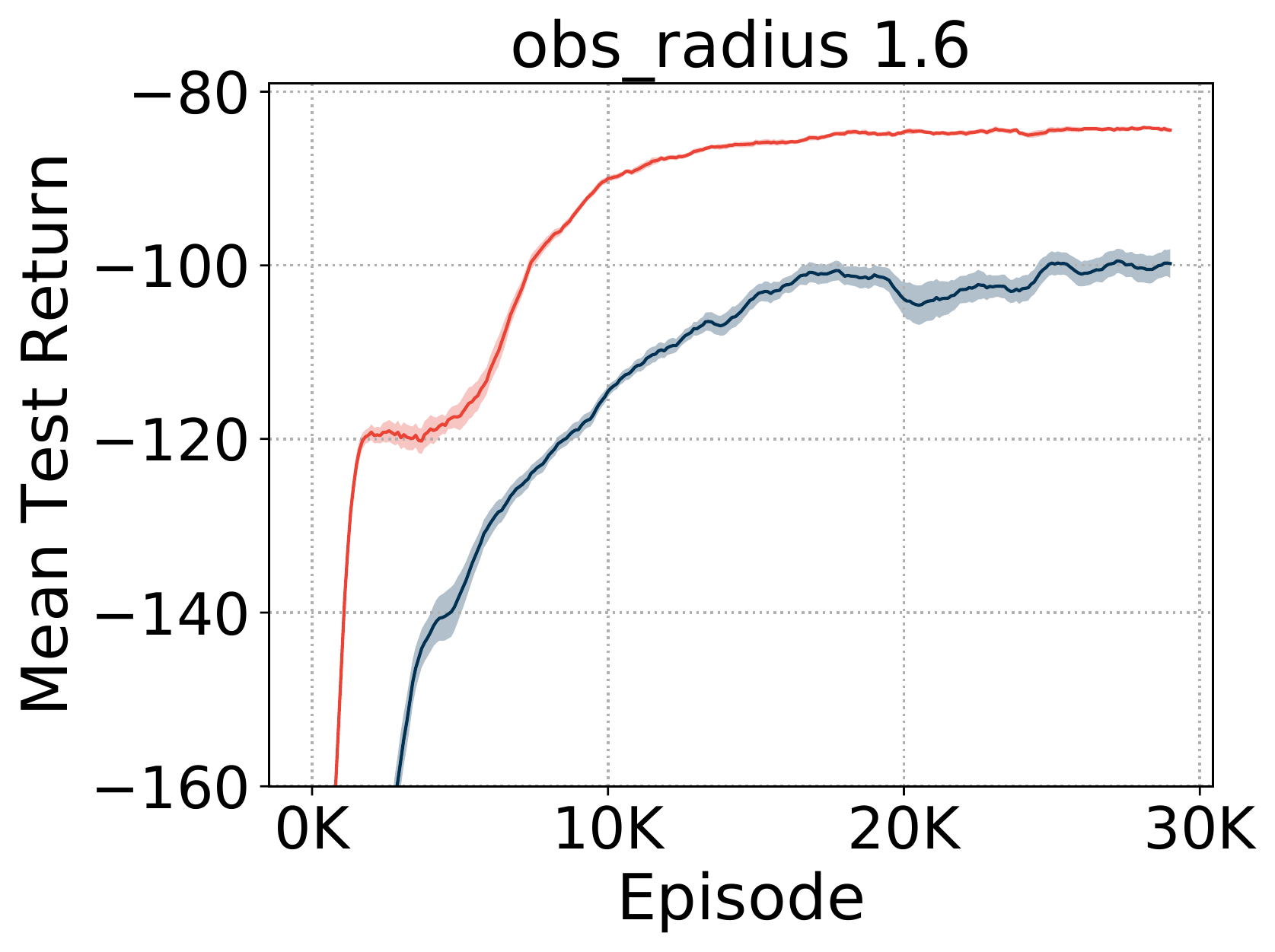}}
    \caption{Comparison against ECA algorithm in two scenarios of each domain.}
    \label{re3}
\end{figure*}

\emph{\textbf{Comparison 3}} (Fig.~\ref{re3}). In the end, we want to validate ROLA's superiorities over the ECA algorithm (explicitly compute the expectation over other agents' joint action space) and highlight the differences between the two approaches.

\subsection{Overall Effectiveness and Robustness of ROLA}

In this section, we provide an analysis of the results shown in Fig.~\ref{re1}.

\textbf{Capture Target} is a classical Dec-POMDP benchmark, in which the transition uncertainties, noisy observation setup and the extremely sparse reward potentially lead to high learning variance~\cite{DecHDRQN}. 
As an on-policy method, under such an environment, ROLA achieves the highest performance with much lower variance on evaluations than Central-V and IA2C (shown in Fig~\ref{re1}a-b) in various grid world sizes, which manifests the significant strength of ROLA on variance reduction. 
As an off-policy learning approach with a replay buffer, SQDDPG is more sample-efficient than other on-policy methods. 
It is why SQDDPG can learn quickly with low variance in the early stage; however, it is troubled with a local optimum as the world space gets large. 
We blame this on the shortcoming of its credit assignment mechanism, depending on \emph{difference rewards}, to deal with such scenarios with sparse reward and unique coalition. 
LIIR certainly suffers from the challenge of approximating intrinsic reward with inadequate external reward. 
In this task, the optimal strategy for each agent is to focus on the target to discern its dynamics while moving towards it, rather than concentrating on the teammate's movements. 
MAAC's failure thus makes sense because it brings extra puzzles to agents by encouraging them to pay attention to each other. 
Additionally, COMA's defects are also exposed in this domain.

In \textbf{Box Pushing}, in order to quickly learn the optimal collaboration, each agent has to reason about its responsibility for a global reward accurately. Otherwise, it is likely to reach a sub-optimum where only one agent pushes a small box to the goal. 
In both the small grid world and the bigger one, ROLA always converges to the optimal values (dash line in Fig.~\ref{re1}c-d) at the fastest speed over other baselines, which proves ROLA's efficiency on credit assignment. 
Furthermore, when the grid world becomes larger, almost all other baselines' performance fluctuations increase while ROLA can keep it relatively low, which shows the advantage of its baseline calculated using the \emph{local critic} on variance reduction. 
Also, IA2C outperforms other centralized baselines because the optimal behaviors require each agent to only focus on pushing the nearest box rather than the other agent's actions, which is also the main reason why both MAAC and SQDDPG perform such poorly due to the attention bias and the strong assumption on coalitions respectively. 
The low efficiency of LIIR and COMA on credit assignment is shown by the slower learning speed even than Central-V.

Unlike the above two domains, \textbf{Cooperative Navigation} generates very dense reward signals throughout the entire episode. At the same time, ROLA still outstands other baselines with a minor performance variance, the fastest convergence speed, and the highest averaged testing return, shown in Fig.~\ref{re1}e-f, which is one solid evidence for ROLA's robustness. 
We also notice that the smaller each agent's view field, the more difficult the task is, and the more pronounced ROLA's advantages will be compared to other baselines.    
This domain requires agents to reach an agreement on the target landmarks allocation. 
IA2C traps at a local optimum due to the lack of information sharing over agents under this partially observable scenario. 
Even having a replay buffer, MAAC learns slowly and tends to converge to a suboptimal solution because it enforces each agent to pay attention to other agents more or less at every step, which is unnecessary for this task and can potentially result in inaccurate credit assignment. 
Same to LIIR, it measures each agent's marginalized contribution over possible coalitions at every time-step, which violates a truth in this domain that agents may only need to coordinate in a few time-steps depending on their positions and landmarks' configuration. 
As the limitations of LIIR and COMA mentioned in the introduction, the shaped reward designed in this domain certainly causes big trouble for LIIR to learn a correct individual reward for each agent and make COMA's counterfactual mechanism out of operation. 
Central-V avoids the sub-optimum due to the less biased underlying uniform credit assignment model.     

\textbf{Antipodal Navigation} is a less cooperative task; this is why SQDDPG performs the worst due to the redundant consideration over all possible coalitions as shown in Fig.~\ref{re1}g-h. 
Thanks to the individual reward function design, LIIR's training improves significantly, and IA2C achieves the second-best performance. 
The inefficiency of COMA's credit assignment design is again affirmed by the phenomenon that both MAAC and COMA run into sub-optimums. 
Even though ROLA's learning is just slightly faster than IA2C and Central-V here, ROLA still performs the best under such a domain where agents have independent goals, demonstrating ROLA's robustness from another degree.   

\subsection{Advantages of ROLA's Learning for Local Critics}

VDAC-sum, VDAC-mix, and DOP all implicitly learn local critics via value decomposition networks. 
VDAC-sum represents the centralized critic as a summation of local critics that severely limits the complexity of the learned centralized value function. 
The local critics in VDAC-sum thus often cannot provide reliable value estimations for policies updates so that VDAC-sum performs the worst in the comparison shown in Fig.~\ref{re2}. 
Despite the fact that the centralized critic of VDAC-mix uses a hyper-network to provide a non-linear mixing of local critics, it is still limited by the monotonicity constraint of the mixing weights learned from global state information. 
Also, the hyper-network takes a longer time to train, so that it delays the decentralized policy optimizations, which explains why VDAC-mix converges slower than ROLA in all considered domains. 
DOP factorizes the centralized critic as a weighted linear summation of individual critics. 
With a replay buffer, DOP's loss for training critics is a mechanism that balances off-policy TD and on-policy TD. It is why DOP outputs better sample efficiency and has less variance than the two versions of VDAC. Nevertheless, DOP's entire performance is still not competitive to ROLA due to the decomposition constraint enforced there.  
Unlike the above VDN-based methods, the centralized critic learned in ROLA approximates the true state-joint-action value function without any constraint. 
We believe, as the results shown in Fig.~\ref{re2}, ROLA's outperformance over the three VDN-based policy gradient algorithms significantly proves the benefit of training local critics via sampling target-actions from the joint Q-value function, which provides necessary data from a centralized perspective to promote learning high-quality decentralized policies. 

\subsection{Superiorities of ROLA over ECA}

The comparison between ROLA and ECA is shown in Fig.~\ref{re3}, where the performance of ROLA exceeds ECA's a lot in terms of both final converged return and sample efficiency under all environments. 
We notice that explicitly calculating the ECA-value using the learned centralized Q-value function actually leads to a strong single approximator bias when the centralized Q-value function has not been fully trained for all agents' joint actions, which is the main reason why ECA performs such poorly.  
Instead, ROLA avoids this bias by learning a global action-value approximator with separate local action-value approximators. 
Moreover, as an on-policy CTDE algorithm, the advantage value estimation in ECA is still with respect to decentralized policies, while ROLA, again, takes extra benefits from the centralized sampling operation. 
In addition, ECA's calculation encounters an exponentially increased computational cost, while ROLA learns each local critic in the new sampling way that is much more practical.

\section{Conclusion}

This paper introduces a new multi-agent actor-critic policy gradient algorithm, named ROLA, to learn decentralized policies with two types of critics. 
ROLA implicitly delivers credit assignment over agents by using each agent's \emph{local critic} that is trained by taking advantage of a centralized critic in a novel sampling process, which enables the CTDE paradigm to utilize action data from centralized selections rather than purely depending on decentralized execution data.
Variance reduction on each agent's decentralized policy gradient estimation is then accomplished by computing local advantage action-values based on the \emph{local critic}. 
The overall robustness and effectiveness of ROLA is empirically established by attaining faster learning, higher returns, and lower variance than a set of state-of-the-art actor-critic methods in diverse multi-agent domains.  

\textbf{Acknowledgements} 
This research is supported in part by the U.S.~Office of Naval
Research under award number N00014-19-1-2131, Army Research
Office award W911NF-20-1-0265 and NSF CAREER Award
2044993.

\clearpage
\bibliographystyle{IEEEtran}
\bibliography{ref}

\clearpage
\onecolumn
\section{Appendix}
\subsection{Algorithm}

We summarize the on-policy version of ROLA in Algorithm~\ref{alg1}, which is the one we use for generating the results in the main paper. The corresponding off-policy learning version can be obtained by disabling line-23 and implementing important sampling weights in line-16, 19, 21

\begin{algorithm}[h]
    \footnotesize
    \caption{Robust Local Advantage (ROLA) Actor-Critic}
    \label{alg1}
        \begin{algorithmic}[1]
            \State Initialize centralized critic networks: $Q_\phi^{\vec{\pi}_\theta}$, $Q_{\phi^-}^{\vec{\pi}_\theta}$
            \State Initialize local critic networks for each agent $i$: $Q_{\psi_i}^{loc}$, $Q_{\psi_i^-}^{loc}$
            \State Initialize decentralized policy networks for each agent $i$: $\pi_{\theta_i}$, $\pi_{\theta^-_i}$
            \State Initialize a joint buffer $\mathcal{B}$
            \For{\emph{episode} = $1$ to $M$}
                \State $t=0$
                \State Reset env
                \While{$s_t$ not terminal \textbf{and} $t < T$}
                    \For{each agent $i$}
                        \State $a_{i,t} \sim \pi_{\theta_i}(o_{i,t},\tau_{i,t-1}; \epsilon)$
                    \EndFor
                    \State Get reward $r_t$, next observations $\vec{o}_{t+1}$, next state $s_{t+1}$
                    \State Collect $\langle s_t, \vec{o}_t,\vec{a}_t, r_t, \vec{o}_{t+1} \rangle$ into buffer $\mathcal{B}$
                    \State $t \leftarrow t + 1$
                \EndWhile
                \If{\emph{episode} mod $I_{\text{train}} == 0$}
                    \State Perform a gradient decent step on $\big(Q_\phi^{\vec{\pi}_\theta}(s, \vec{a})-y\big)^2_\mathcal{B}$ 
                    \State $y = r + \gamma Q_{\phi^-}^{\vec{\pi}_{\theta^-}}(s, a_1',...,a_n') \mid _{a_j'\sim\pi_{\theta_j^-}(\tau_j')}$

                    \For{1 ... num local critic update}
                        \State Perform a gradient decent step on $\big(Q_{\psi_i}^{loc}(s, a_i) - y\big)^2_\mathcal{B}$
                        \State $y = r + \gamma Q_{\psi_i^-}^{loc}(s, a_i') \mid _{a_i'\leftarrow \vec{a}\,' \sim \text{softmax} \big(Q_\phi^{\vec{\pi}_{\theta}}(s',\vec{a}\,')\big)}$
 
                    \EndFor
                    \For{each agent i}
                        \State Perform a gradient ascent on $\nabla_{\theta_i} J(\theta_i) = \mathbb{E}_{\vec{\pi}_\theta}\biggr[\nabla_{\theta_i}\log\pi_{\theta_i}(a_i|\tau_i)A_i(s,a_i)\biggr]$
                        \State $A_i(s, a_i) = Q_{\psi_i}^{loc}(s, a_i) - \underset{\dot{a_i}}{\sum}\pi_{\theta_i}(\dot{ a_i }|\tau_i)Q_{\psi_i}^{loc}(s,\dot{a_i})$ 
                    \EndFor
                    \State Reset buffer $\mathcal{B}$
                \EndIf
                \If{\emph{episode} mod $I_{\text{TargetUpdate}} = 0$}
                    \State Update centralized critic target network $\phi^-\leftarrow\phi$
                    \State Update each agent $i$'s local critic target network $\psi_i^-\leftarrow\psi_i$
                    \State Update each agent $i$'s decentralized policy target network $\theta_i^-\leftarrow\theta_i$
                \EndIf
            \EndFor
        \end{algorithmic}
\end{algorithm}

\clearpage
\subsection{Domain Details}

\emph{\textbf{Capture Target.}} 

\begin{figure}[h!]
    \centering
    \includegraphics[height=3.5cm]{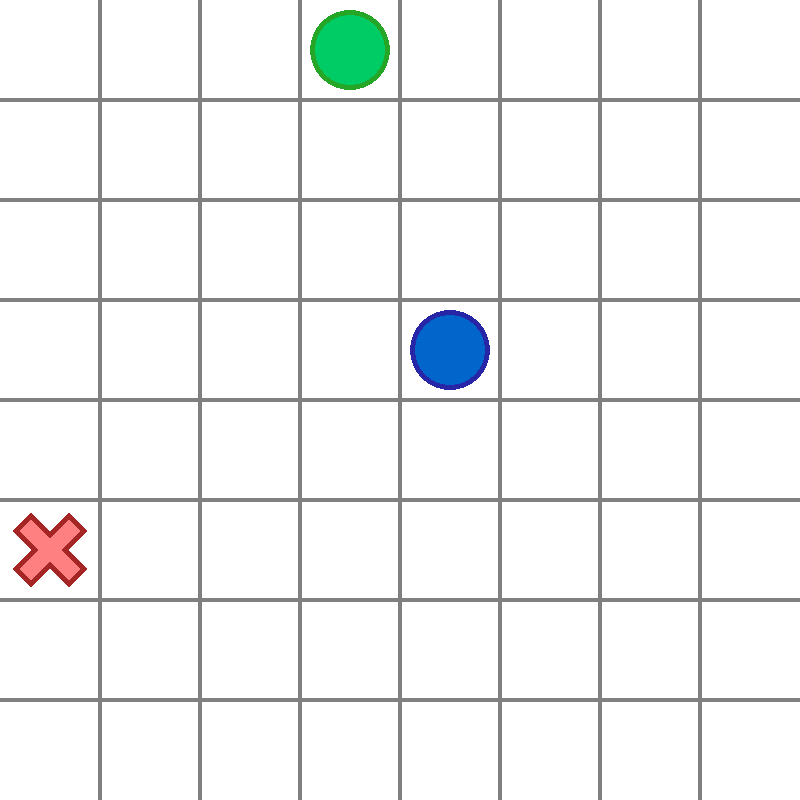}
\end{figure}

\textbf{Goal}: Two agents move in a $n\times n$ toroidal grid world to capture a moving target by simultaneously arriving same grid cell as where the target is. The positions of agents and the target are randomly initialized.   

\textbf{State:} The global state information involve the positions of each agent and the target under a gird world. 

\textbf{Action Space:} Each agent has five applicable actions: \emph{up}, \emph{down}, \emph{left}, \emph{right} and \emph{stay}.

\textbf{Observation Space:} Each agent can always observe its own location (not teammate's) and sometimes observe the target's location with a probability 0.7.

\textbf{Dynamics:} The transition noise for each agent is a probability 0.1 of accidentally locating at a random adjacent cell, which the target deterministically  moves towards east. 

\textbf{Reward:} Agents can only receive a terminal reward $+1$ when they successfully capture the target.

\textbf{Termination:} Each episode terminates either the target captured or after 60 time-steps.\\
\\
\\
\\
\\
\\
\\

\emph{\textbf{Small Box Pushing.}}

\begin{figure}[h!]
    \centering
    \includegraphics[height=3.5cm]{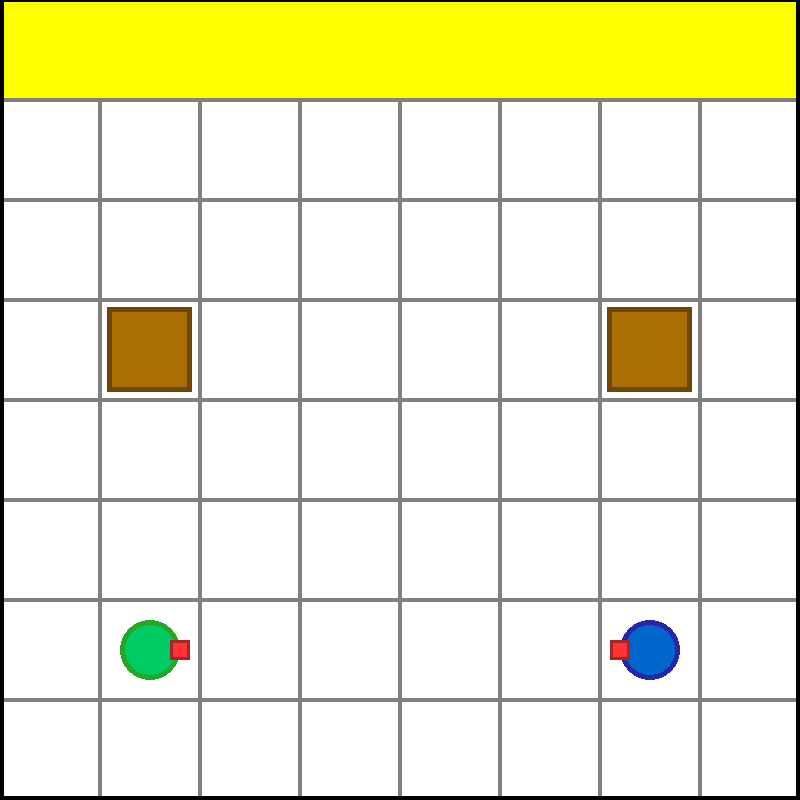}
\end{figure}

\textbf{Goal:} Two agents are tasked with pushing two boxes to the goal area at the top of a grid world. Agents and boxes are initialized deterministically.  

\textbf{State:} The global state information consist of the each agent's location and each box's position in a grid world. 

\textbf{Action Space:} Each agent has four applicable actions: \emph{move forward}, \emph{turn left}, \emph{turn right}, and \emph{stay}.

\textbf{Observation Space:} Each agent is only allowed to capture the front cell's state: \emph{empty}, \emph{box}, \emph{teammate}, or \emph{boundary}.

\textbf{Dynamics:} The transition in this task is deterministic. The box is only allowed to be moved towards north when any agent faces it at the south side and executes \emph{move forward}.   

\textbf{Reward:} When any box is pushed to the goal area, the team receive a reward $+100$.

\textbf{Termination:} Each episode terminates when any box is pushed to the goal area, or after 100 time-steps.\\

\clearpage

\emph{\textbf{Cooperative Navigation.}}

\begin{figure}[h!]
    \centering
    \includegraphics[height=3.5cm]{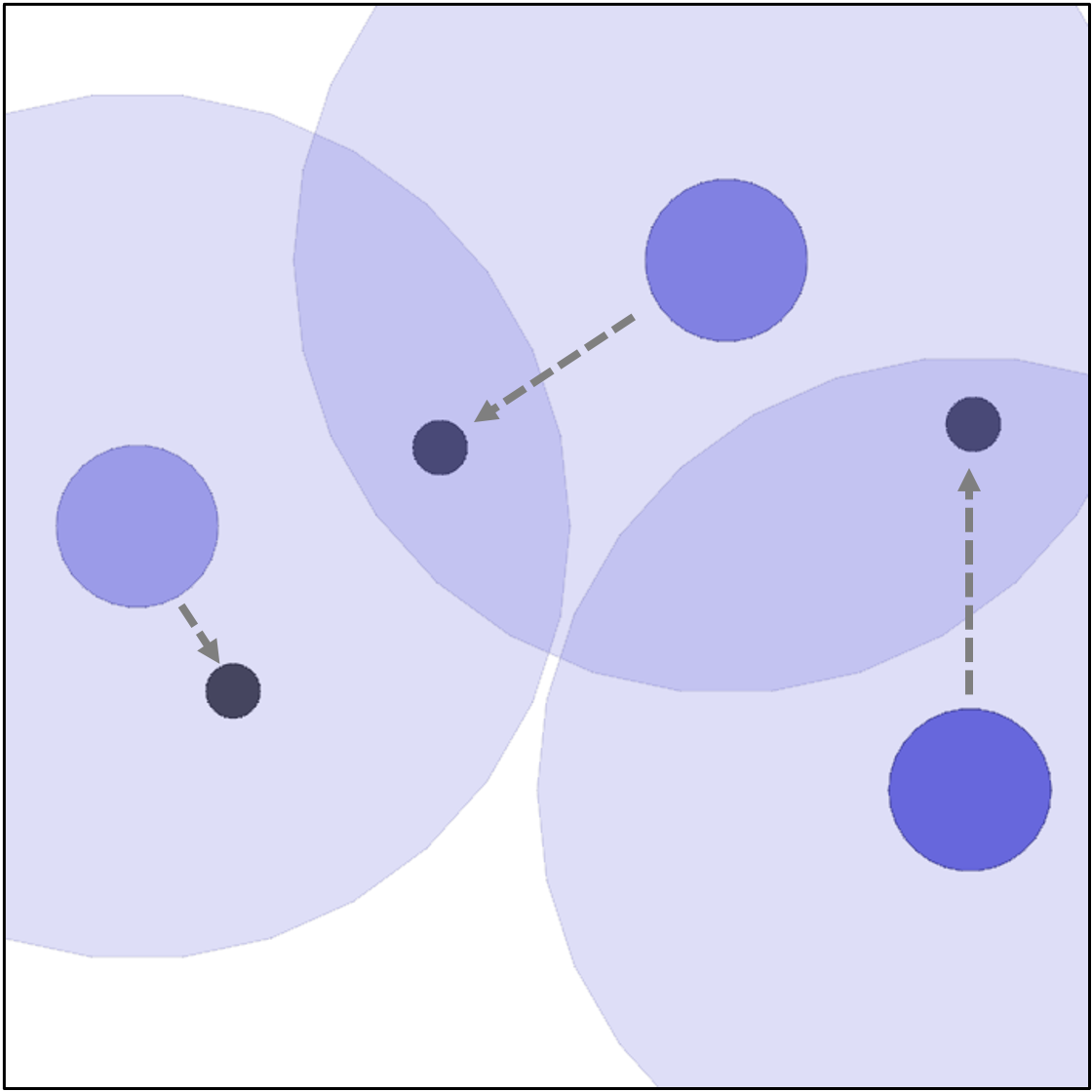}
\end{figure}

\textbf{Goal:} Three agents move around in a unbounded continuous 2-D space and target on covering three randomly initialized static landmarks while avoid colliding with each other. Agents and landmarks are Initialized within the space $(-1,1)^2$. 

\textbf{State:} The global state information includes each agent's absolute 2-D position $(x,y)\in \mathbb{R}^2$ and velocity $(v_x, v_y) \in \mathbb{R}^2$. 

\textbf{Action Space:} Agent's movement is achieved by applying forces on the agent's physical body. There are 9 discrete force vectors: $[0,0]$, $[1,0]$, $[1,-1]$, $[-1,-1]$, $[-1,0]$, $[-1,-1]$, $[-1,0]$, $[-1,1]$ and $[0,1]$.   

\textbf{Observation Space:} Each agent's observation includes: its own absolute position and velocity, landmarks' absolute positions and other agents' relative positions and velocities with respect to itself in side of a pre-defined view field. 

\textbf{Dynamics:} A simple physics model involves inertia and friction effects. Contact force also exists when collision happens. 

\textbf{Reward:} At each time step, agents obtain a global reward that is the negative summation of the distance from each landmark to the corresponding closest agent. 

\textbf{Termination:} Each episode terminates after 25 time-steps.\\
\\
\\
\\
\\

\emph{\textbf{Antipodal Navigation.}}

\begin{figure}[h!]
    \centering
    \includegraphics[height=3.5cm]{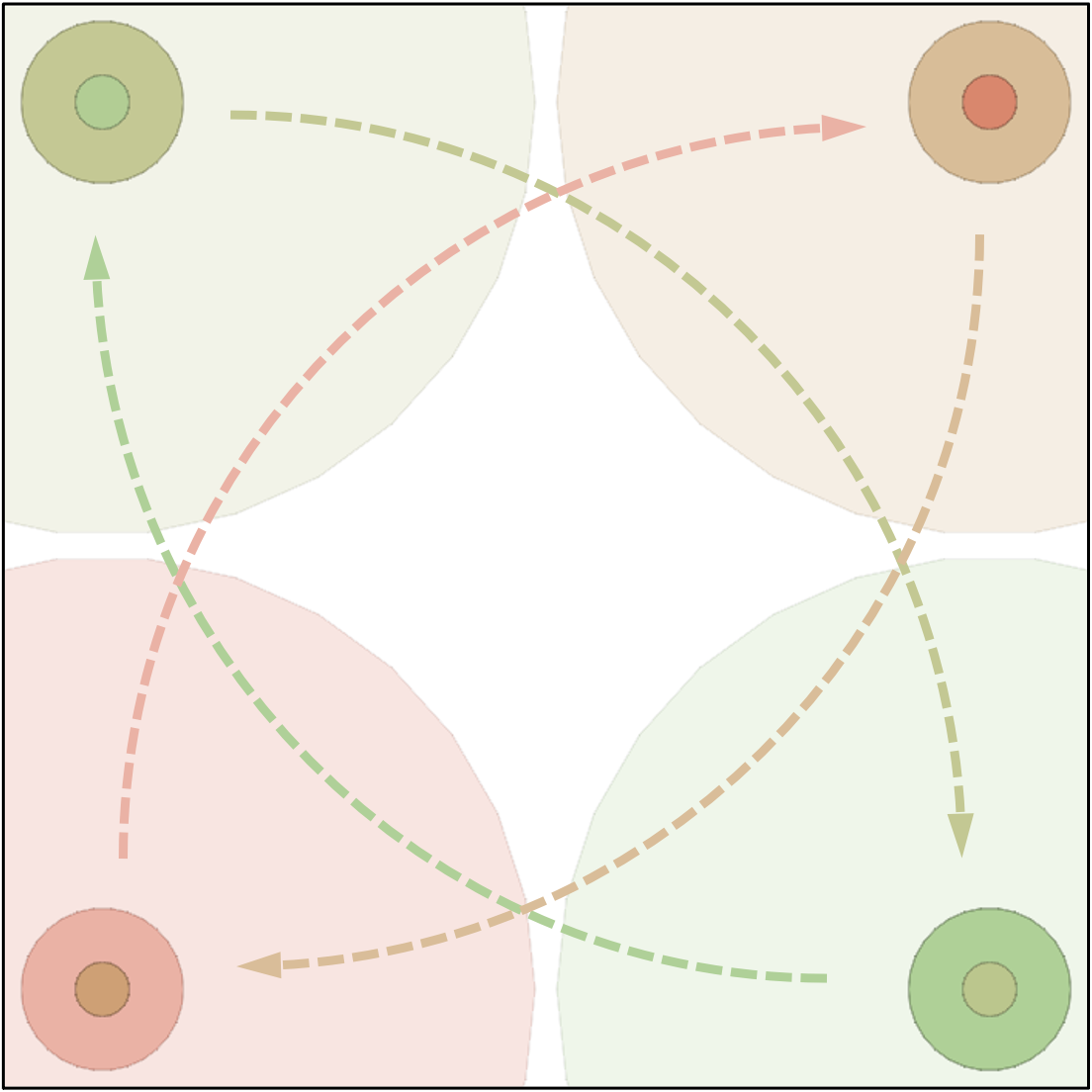}
\end{figure}

\textbf{Goal:} Four agents move in an unbounded continuous 2-D space with the objective to cover each own allocated target landmark. With probability 0.2, the initial positions of landmarks and agents are uniformly drawn in the 2-D space $(-1,1)^2$, otherwise their positions are initialized in an antipodal configuration shown above.  

\textbf{State:} The global state information includes each agent's absolute 2-D position $(x,y)\in \mathbb{R}^2$ and velocity $(v_x, v_y) \in \mathbb{R}^2$. 

\textbf{Action Space:} Agent's movement is achieved by applying forces on the agent's physical body. There are 5 discrete force vectors: $[0,0]$, $[1,0]$, $[-1,0]$, $[0,-1]$ and $[0,1]$.   

\textbf{Observation Space:} Each agent's observation includes: its own absolute position and velocity, landmarks' absolute positions and other agents' relative positions and velocities with respect to itself in side of a pre-defined view field. 

\textbf{Dynamics:} A simple physics model involves inertia and friction effects. Contact force also exists when collision happens. 

\textbf{Reward:} At every time-step, an individual reward is assigned to each agent, which is the negative distance between the agent and its target landmark. Besides, a penalty $-1$ is issued to the corresponding agent when it collides with other agents. 

\textbf{Termination}. Each episode terminates when agents' distances to their targets are all less than 0.05, or after 50 time steps.

\clearpage
\subsection{Hyper-parameter Summarization}

\begin{table}[h!]
    \caption {Hyper-parameters used for methods achieving the best performance in Capture Target 6x6 grid world.}
    \centering
    \begin{tabular}{lcccccccccc}
    \toprule
        Parameter & ROLA & DOP & VDAC-mix & VDAC-sum & MAAC & SQDDPG & LIIR & COMA & Central-V & IA2C\\
    \cmidrule(r){2-11}
        Training Episodes & 1e5 & 1e5 & 1e5 & 1e5 & 1e5 & 1e5 & 1e5 & 1e5 & 1e5 & 1e5 \\
        Actor learning rate & 5e-4 & 5e-4 & 5e-4 & 3e-4 & 1e3 & 5e-4 & 5e-4 & 5e-4 & 3e-4 & 5e-4\\
        Critic learning rate & 5e-4 & 5e-4 & 5e-4 & 3e-3 & 1e-3 & 5e-4 & 5e-4 & 1e-3 & 3e-3 & 5e-4\\
        Episodes per train & 2 & 2 & 2 & 2 & N/A & N/A & 8 & 8 & 8 & 2 \\
        Target-net update freq (episode) & 16 & 32 & 64 & 32 & N/A & 200 (step) & 32 & 32 & 16 & 32\\
        N-step TD & 3 & N/A & N/A & N/A & 1 & 1 & N/A & N/A & 1 & 1 \\
        Num centralized critic update & 1 & 1 & 1 & 1 & 4 & 4 & 1 & 1 & 1 & N/A\\
        Num local critic update & 1 & 1 & 1 & 1 & N/A & N/A & N/A & N/A & N/A & 1\\
        $\epsilon_{\text{start}}$ & 1.0 & 1.0 & 1.0 & 1.0 & N/A & N/A & 1.0 & 1.0 & 1.0 & 1.0\\
        $\epsilon_{\text{end}}$ & 0.05 & 0.05 & 0.05 & 0.05 & N/A & N/A & 0.05 & 0.05 & 0.05 & 0.05 \\
        $\epsilon_{\text{decay}}$ (episode) & 15e3 & 15e3 & 15e3 & 15e3 & N/A & N/A & 15e3 & 15e3 & 15e3 & 15e3 \\
        TD$(\lambda)$ & N/A & 0.3 & 0.8 & 0.8 & N/A & N/A & 0.3 & 0.3 & N/A & N/A \\
        Steps per train & N/A & N/A & N/A & N/A & 25 & 25 & N/A & N/A & N/A & N/A\\
        Target-net soft-update rate  & N/A & N/A & N/A & N/A & 5e-3 & 1e-2 & N/A & N/A & N/A & N/A\\
        Entropy loss weight & N/A & N/A & N/A & N/A & 1e-4 & 1e-4 & N/A & N/A & N/A & N/A\\
        Replay buffer size & N/A & 5000 & N/A & N/A & 5e3 & 5e3 & N/A & N/A & N/A & N/A \\ 
        Batch size (episode) & N/A & 32 & N/A & N/A & 4 & 4 & N/A & N/A & N/A & N/A  \\ 
        Num attention head & N/A & N/A & N/A & N/A & 1 & N/A & N/A & N/A & N/A & N/A  \\
    \bottomrule
    \end{tabular} 
\end{table}

\begin{table}[h!]
    \caption {Hyper-parameters used for methods achieving the best performance in Capture Target 8x8 grid world.}
    \centering
    \begin{tabular}{lcccccccccc}
    \toprule
        Parameter & ROLA & DOP & VDAC-mix & VDAC-sum & MAAC & SQDDPG & LIIR & COMA & Central-V & IA2C\\
    \cmidrule(r){2-11}
        Training Episodes & 2e5 & 2e5 & 2e5 & 2e5 & 2e5 & 2e5 & 2e5 & 2e5 & 2e5 & 2e5 \\
        Actor learning rate & 5e-4 & 5e-4 & 3e-4 & 5e-4 & 1e-3 & 5e-4 & 5e-4 & 5e-4 & 3e-4 & 5e-4\\
        Critic learning rate & 5e-4 & 5e-4 & 3e-3 & 5e-4 & 1e-3 & 5e-4 & 1e-3 & 1e-3 & 3e-3 & 5e-4\\
        Episodes per train & 2 & 8 & 2 & 2 & N/A & N/A & 8 & 8 & 8 & 8 \\
        Target-net update freq (episode) & 64 & 64 & 64 & 32 & N/A & 200 (step) & 16 & 64 & 16 & 64\\
        N-step TD & 3 & N/A & N/A & N/A & 1 & 1 & N/A & N/A & 1 & 1 \\
        Num centralized critic update & 1 & 1 & 1 & 1 & 4 & 4 & 1 & 1 & 1 & N/A\\
        Num local critic update & 1 & 1 & 1 & 1 & N/A & N/A & N/A & N/A & N/A & 1\\
        $\epsilon_{\text{start}}$ & 1.0 & 1.0 & 1.0 & 1.0 & N/A & N/A & 1.0 & 1.0 & 1.0 & 1.0\\
        $\epsilon_{\text{end}}$ & 0.05 & 0.05 & 0.05 & 0.05 & N/A & N/A & 0.05 & 0.05 & 0.05 & 0.05 \\
        $\epsilon_{\text{decay}}$ (episode) & 15e3 & 15e3 & 15e3 & 15e3 & N/A & N/A & 15e3 & 15e3 & 15e3 & 15e3 \\
        TD$(\lambda)$ & N/A & 0.3 & 0.3 & 0.8 & N/A & N/A & 0.3 & 0.3 & N/A & N/A \\
        Steps per train & N/A & N/A & N/A & N/A & 50 & 25 & N/A & N/A & N/A & N/A\\
        Target-net soft-update rate  & N/A & N/A & N/A & N/A & 5e-3 & 1e-2 & N/A & N/A & N/A & N/A\\
        Entropy loss weight & N/A & N/A & N/A & N/A & 1e-4 & 1e-4 & N/A & N/A & N/A & N/A\\
        Replay buffer size & N/A & 5000 & N/A & N/A & 5e3 & 5e3 & N/A & N/A & N/A & N/A \\ 
        Batch size (episode) & N/A & 32 & N/A & N/A & 4 & 4 & N/A & N/A & N/A & N/A  \\ 
        Num attention head & N/A & N/A & N/A & N/A & 1 & N/A & N/A & N/A & N/A & N/A  \\
    \bottomrule
    \end{tabular} 
\end{table}

\begin{table}[h!]
    \caption {Hyper-parameters used for methods achieving the best performance in Box Pushing 6x6 grid world.}
    \centering
    \begin{tabular}{lcccccccccc}
    \toprule
        Parameter & ROLA & DOP & VDAC-mix & VDAC-sum & MAAC & SQDDPG & LIIR & COMA & Central-V & IA2C\\
    \cmidrule(r){2-11}
        Training Episodes & 4e3 & 4e3 & 4e3 & 4e3 & 4e3 & 4e3 & 4e3 & 4e3 & 4e3 & 4e3 \\
        Actor learning rate & 1e-3 & 1e-3 & 1e-3 & 1e-3 & 5e-4 & 1e-3 & 1e-3 & 1e-3 & 1e-3 & 1e-3\\
        Critic learning rate & 3e-3 & 1e-3 & 1e-3 & 1e-3 & 1e-3 & 1e-3 & 5e-3 & 3e-3 & 5e-3 & 5e-3\\
        Episodes per train & 2 & 2 & 4 & 2 & N/A & N/A & 4 & 8 & 2 & 2 \\
        Target-net update freq (episode) & 32 & 32 & 32 & 32 & N/A & N/A & 64 & 16 & 64 & 32\\
        N-step TD & 3 & N/A & N/A & N/A & 1 & 1 & N/A & N/A & 3 & 5 \\
        Num centralized critic update & 1 & 1 & 1 & 1 & 4 & 4 & 1 & 1 & 1 & N/A\\
        Num local critic update & 4 & 1 & 1 & 1 & N/A & N/A & N/A & N/A & N/A & 1\\
        $\epsilon_{\text{start}}$ & 1.0 & 1.0 & 1.0 & 1.0 & N/A & N/A & 1.0 & 1.0 & 1.0 & 1.0\\
        $\epsilon_{\text{end}}$ & 0.01 & 0.01 & 0.01 & 0.01 & N/A & N/A & 0.01 & 0.01 & 0.01 & 0.01 \\
        $\epsilon_{\text{decay}}$ (episode) & 2e3 & 2e3 & 2e3 & 2e3 & N/A & N/A & 2e3 & 2e3 & 2e3 & 2e3 \\
        TD$(\lambda)$ & N/A & 0.8 & 0.4 & 0.4 & N/A & N/A & 0.8 & 0.4 & N/A & N/A \\
        Steps per train & N/A & N/A & N/A & N/A & 25 & 25 & N/A & N/A & N/A & N/A\\
        Target-net soft-update rate  & N/A & N/A & N/A & N/A & 5e-3 & 1e-1 & N/A & N/A & N/A & N/A\\
        Entropy loss weight & N/A & N/A & N/A & N/A & 1e-2 & 1e-3 & N/A & N/A & N/A & N/A\\
        Replay buffer size & N/A & 500 & N/A & N/A & 2e3 & 5e2 & N/A & N/A & N/A & N/A \\ 
        Batch size (episode) & N/A & 32 & N/A & N/A & 32 & 16 & N/A & N/A & N/A & N/A  \\ 
        Num attention head & N/A & N/A & N/A & N/A & 1 & N/A & N/A & N/A & N/A & N/A  \\
    \bottomrule
    \end{tabular} 
\end{table}

\begin{table}[h!]
    \caption {Hyper-parameters used for methods achieving the best performance in Box Pushing 10x10 grid world.}
    \centering
    \begin{tabular}{lcccccccccc}
    \toprule
        Parameter & ROLA & DOP & VDAC-mix & VDAC-sum & MAAC & SQDDPG & LIIR & COMA & Central-V & IA2C\\
    \cmidrule(r){2-11}
        Training Episodes & 4e3 & 4e3 & 4e3 & 4e3 & 4e3 & 4e3 & 4e3 & 4e3 & 4e3 & 4e3 \\
        Actor learning rate & 5e-4 & 1e-3 & 1e-3 & 1e-4 & 1e-4 & 5e-4 & 1e-3 & 3e-4 & 5e-4 & 1e-3\\
        Critic learning rate & 1e-3 & 1e-3 & 1e-3 & 1e-3 & 1e-3 & 5e-4 & 5e-3 & 3e-3 & 5e-4 & 3e-3\\
        Episodes per train & 2 & 2 & 2 & 2 & N/A & N/A & 8 & 8 & 4 & 2 \\
        Target-net update freq (episode) & 16 & 16 & 16 & 32 & N/A & N/A & 64 & 16 & 16 & 64\\
        N-step TD & 1 & N/A & N/A & N/A & 1 & 1 & N/A & N/A & 1 & 5 \\
        Num centralized critic update & 1 & 1 & 1 & 1 & 4 & 4 & 1 & 1 & 1 & N/A\\
        Num local critic update & 4 & 1 & 1 & 1 & N/A & N/A & N/A & N/A & N/A & 1\\
        $\epsilon_{\text{start}}$ & 1.0 & 1.0 & 1.0 & 1.0 & N/A & N/A & 1.0 & 1.0 & 1.0 & 1.0\\
        $\epsilon_{\text{end}}$ & 0.01 & 0.01 & 0.01 & 0.01 & N/A & N/A & 0.01 & 0.01 & 0.01 & 0.01 \\
        $\epsilon_{\text{decay}}$ (episode) & 4e3 & 4e3 & 4e3 & 4e3 & N/A & N/A & 4e3 & 4e3 & 4e3 & 4e3 \\
        TD$(\lambda)$ & N/A & 0.8 & 0.4 & 0.4 & N/A & N/A & 0.8 & 0.4 & N/A & N/A \\
        Steps per train & N/A & N/A & N/A & N/A & 50 & 25 & N/A & N/A & N/A & N/A\\
        Target-net soft-update rate  & N/A & N/A & N/A & N/A & 5e-3 & 1e-1 & N/A & N/A & N/A & N/A\\
        Entropy loss weight & N/A & N/A & N/A & N/A & 1e-2 & 1e-3 & N/A & N/A & N/A & N/A\\
        Replay buffer size & N/A & 500 & N/A & N/A & 3e3 & 5e2 & N/A & N/A & N/A & N/A \\ 
        Batch size (episode) & N/A & 32 & N/A & N/A & 16 & 16 & N/A & N/A & N/A & N/A  \\ 
        Num attention head & N/A & N/A & N/A & N/A & 1 & N/A & N/A & N/A & N/A & N/A  \\
    \bottomrule
    \end{tabular} 
\end{table}

\begin{table}[h!]
    \caption {Hyper-parameters used for methods achieving the best performance in Cooperative Navigation with observation radius 1.6.}
    \centering
    \begin{tabular}{lcccccccccc}
    \toprule
        Parameter & ROLA & DOP & VDAC-mix & VDAC-sum & MAAC & SQDDPG & LIIR & COMA & Central-V & IA2C\\
    \cmidrule(r){2-11}
        Training Episodes & 1e5 & 1e5 & 1e5 & 1e5 & 1e5 & 1e5 & 1e5 & 1e5 & 1e5 & 1e5 \\
        Actor learning rate & 1e-3 & 1e-3 & 1e-3 & 1e-3 & 1e-3 & 1e-3 & 5e-4 & 5e-4 & 1e-3 & 5e-4\\
        Critic learning rate & 1e-3 & 5e-3 & 5e-3 & 5e-3 & 1e-3 & 1e-3 & 5e-4 & 1e-3 & 1e-3 & 3e-3\\
        Episodes per train & 2 & 2 & 4 & 4 & N/A & N/A & 2 & 2 & 2 & 2 \\
        Target-net update freq (episode) & 16 & 32 & 32 & 16 & N/A & N/A & 16 & 8 & 8 & 8\\
        N-step TD & 5 & N/A & N/A & N/A & 1 & 1 & N/A & N/A & 3 & 5 \\
        Num centralized critic update & 1 & 1 & 1 & 1 & 4 & 4 & 1 & 1 & 1 & N/A\\
        Num local critic update & 4 & 1 & 1 & 1 & N/A & N/A & N/A & N/A & N/A & 1\\
        $\epsilon_{\text{start}}$ & 1.0 & 1.0 & 1.0 & 1.0 & N/A & N/A & 1.0 & 1.0 & 1.0 & 1.0\\
        $\epsilon_{\text{end}}$ & 0.05 & 0.05 & 0.05 & 0.05 & N/A & N/A & 0.05 & 0.05 & 0.05 & 0.05 \\
        $\epsilon_{\text{decay}}$ (episode) & 5e4 & 5e4 & 5e4 & 5e4 & N/A & N/A & 5e4 & 5e4 & 5e4 & 5e4 \\
        TD$(\lambda)$ & N/A & 0.3 & 0.8 & 0.8 & N/A & N/A & 0.3 & 0.3 & N/A & N/A \\
        Steps per train & N/A & N/A & N/A & N/A & 100 & 100 & N/A & N/A & N/A & N/A\\
        Target-net soft-update rate & N/A & N/A & N/A & N/A & 5e-3 & 1e-1 & N/A & N/A & N/A & N/A\\
        Entropy loss weight & N/A & N/A & N/A & N/A & 1e-2 & 1e-3 & N/A & N/A & N/A & N/A\\
        Replay buffer size & N/A & 5000 & N/A & N/A & 1e4 & 1e4 & N/A & N/A & N/A & N/A \\ 
        Batch size (episode) & N/A & 32 & N/A & N/A & 32 & 8 & N/A & N/A & N/A & N/A  \\ 
        Num attention head & N/A & N/A & N/A & N/A & 4 & N/A & N/A & N/A & N/A & N/A  \\
    \bottomrule
    \end{tabular} 
\end{table}

\begin{table}[h!]
    \caption {Hyper-parameters used for methods achieving the best performance in Cooperative Navigation with observation radius 1.4.}
    \centering
    \begin{tabular}{lcccccccccc}
    \toprule
        Parameter & ROLA & DOP & VDAC-mix & VDAC-sum & MAAC & SQDDPG & LIIR & COMA & Central-V & IA2C\\
    \cmidrule(r){2-11}
        Training Episodes & 1e5 & 1e5 & 1e5 & 1e5 & 1e5 & 1e5 & 1e5 & 1e5 & 1e5 & 1e5 \\
        Actor learning rate & 1e-3 & 1e-3 & 1e-3 & 5e-4 & 1e-4 & 1e-3 & 5e-4 & 5e-4 & 5e-4 & 3e-4\\
        Critic learning rate & 1e-3 & 5e-3 & 5e-3 & 3e-3 & 1e-3 & 1e-3 & 5e-4 & 1e-3 & 3e-3 & 3e-3\\
        Episodes per train & 2 & 2 & 2 & 4 & N/A & N/A & 2 & 2 & 2 & 2 \\
        Target-net update freq (episode) & 8 & 8 & 16 & 16 & N/A & N/A & 16 & 8 & 16 & 16\\
        N-step TD & 5 & N/A & N/A & N/A & 1 & 1 & N/A & N/A & 3 & 5 \\
        Num centralized critic update & 1 & 1 & 1 & 1 & 4 & 4 & 1 & 1 & 1 & N/A\\
        Num local critic update & 4 & 1 & 1 & 1 & N/A & N/A & N/A & N/A & N/A & 1\\
        $\epsilon_{\text{start}}$ & 1.0 & 1.0 & 1.0 & 1.0 & N/A & N/A & 1.0 & 1.0 & 1.0 & 1.0\\
        $\epsilon_{\text{end}}$ & 0.05 & 0.05 & 0.05 & 0.05 & N/A & N/A & 0.05 & 0.05 & 0.05 & 0.05 \\
        $\epsilon_{\text{decay}}$ (episode) & 5e4 & 5e4 & 5e4 & 5e4 & N/A & N/A & 5e4 & 5e4 & 5e4 & 5e4 \\
        TD$(\lambda)$ & N/A & 0.3 & 0.3 & 0.8 & N/A & N/A & 0.3 & 0.3 & N/A & N/A \\
        Steps per train & N/A & N/A & N/A & N/A & 100 & 100 & N/A & N/A & N/A & N/A\\
        Target-net soft-update rate  & N/A & N/A & N/A & N/A & 5e-3 & 1e-2 & N/A & N/A & N/A & N/A\\
        Entropy loss weight & N/A & N/A & N/A & N/A & 1e-3 & 1e-3 & N/A & N/A & N/A & N/A\\
        Replay buffer size & N/A & 5000 & N/A & N/A & 1e4 & 1e4 & N/A & N/A & N/A & N/A \\ 
        Batch size (episode) & N/A & 32 & N/A & N/A & 32 & 8 & N/A & N/A & N/A & N/A  \\ 
        Num attention head & N/A & N/A & N/A & N/A & 4 & N/A & N/A & N/A & N/A & N/A  \\
    \bottomrule
    \end{tabular} 
\end{table}

\begin{table}[h!]
    \caption {Hyper-parameters used for methods achieving the best performance in Antipodal Navigation with observation radius 1.6.}
    \centering
    \begin{tabular}{lcccccccccc}
    \toprule
        Parameter & ROLA & DOP & VDAC-mix & VDAC-sum & MAAC & SQDDPG & LIIR & COMA & Central-V & IA2C\\
    \cmidrule(r){2-11}
        Training Episodes & 3e4 & 3e4 & 3e4 & 3e4 & 3e4 & 3e4 & 3e4 & 3e4 & 3e4 & 3e4 \\
        Actor learning rate & 5e-4 & 1e-3 & 5e-4 & 5e-4 & 1e-3 & 5e-4 & 5e-4 & 5e-4 & 3e-4 & 3e-4\\
        Critic learning rate & 1e-3 & 1e-3 & 1e-3 & 1e-3 & 1e-3 & 5e-4 & 5e-4 & 5e-4 & 3e-3 & 3e-3\\
        Episodes per train & 2 & 2 & 2 & 8 & N/A & N/A & 2 & 2 & 2 & 2 \\
        Target-net update freq (episode) & 8 & 32 & 8 & 8 & N/A & N/A & 16 & 8 & 8 & 16\\
        N-step TD & 5 & N/A & N/A & N/A & 1 & 1 & N/A & N/A & 5 & 5 \\
        Num centralized critic update & 1 & 1 & 1 & 1 & 4 & 4 & 1 & 1 & 1 & N/A\\
        Num local critic update & 4 & 1 & 1 & 1 & N/A & N/A & N/A & N/A & N/A & 1\\
        $\epsilon_{\text{start}}$ & 1.0 & 1.0 & 1.0 & 1.0 & N/A & N/A & 1.0 & 1.0 & 1.0 & 1.0\\
        $\epsilon_{\text{end}}$ & 0.05 & 0.05 & 0.05 & 0.05 & N/A & N/A & 0.05 & 0.05 & 0.05 & 0.05 \\
        $\epsilon_{\text{decay}}$ (episode) & 2e4 & 2e4 & 2e4 & 2e4 & N/A & N/A & 2e4 & 2e4 & 2e4 & 2e4 \\
        TD$(\lambda)$ & N/A & 0.6 & 0.8 & 0.8 & N/A & N/A & 0.6 & 0.8 & N/A & N/A \\
        Steps per train & N/A & N/A & N/A & N/A & 100 & 100 & N/A & N/A & N/A & N/A\\
        Target-net soft-update rate  & N/A & N/A & N/A & N/A & 5e-3 & 1e-1 & N/A & N/A & N/A & N/A\\
        Entropy loss weight & N/A & N/A & N/A & N/A & 1e-3 & 1e-2 & N/A & N/A & N/A & N/A\\
        Replay buffer size & N/A & 5000 & N/A & N/A & 1e4 & 1e4 & N/A & N/A & N/A & N/A \\ 
        Batch size (episode) & N/A & 32 & N/A & N/A & 2 & 8 & N/A & N/A & N/A & N/A  \\ 
        Num attention head & N/A & N/A & N/A & N/A & 4 & N/A & N/A & N/A & N/A & N/A  \\
    \bottomrule
    \end{tabular} 
\end{table}

\begin{table}[h!]
    \caption {Hyper-parameters used for methods achieving the best performance in Antipodal Navigation with observation radius 1.4.}
    \centering
    \begin{tabular}{lcccccccccc}
    \toprule
        Parameter & ROLA & DOP & VDAC-mix & VDAC-sum & MAAC & SQDDPG & LIIR & COMA & Central-V & IA2C\\
    \cmidrule(r){2-11}
        Training Episodes & 3e4 & 3e4 & 3e4 & 3e4 & 3e4 & 3e4 & 3e4 & 3e4 & 3e4 & 3e4 \\
        Actor learning rate & 5e-4 & 1e-3 & 5e-4 & 5e-4 & 1e-3 & 5e-4 & 5e-4 & 5e-4 & 3e-4 & 3e-4\\
        Critic learning rate & 1e-3 & 1e-3 & 3e-3 & 1e-3 & 1e-3 & 5e-4 & 5e-4 & 5e-4 & 3e-3 & 3e-3\\
        Episodes per train & 2 & 2 & 2 & 8 & N/A & N/A & 2 & 2 & 2 & 2 \\
        Target-net update freq (episode) & 16 & 16 & 32 & 32 & N/A & N/A & 16 & 8 & 8 & 16\\
        N-step TD & 5 & N/A & N/A & N/A & 1 & 1 & N/A & N/A & 5 & 5 \\
        Num centralized critic update & 1 & 1 & 1 & 1 & 4 & 4 & 1 & 1 & 1 & N/A\\
        Num local critic update & 4 & 1 & 1 & 1 & N/A & N/A & N/A & N/A & N/A & 1\\
        $\epsilon_{\text{start}}$ & 1.0 & 1.0 & 1.0 & 1.0 & N/A & N/A & 1.0 & 1.0 & 1.0 & 1.0\\
        $\epsilon_{\text{end}}$ & 0.05 & 0.05 & 0.05 & 0.05 & N/A & N/A & 0.05 & 0.05 & 0.05 & 0.05 \\
        $\epsilon_{\text{decay}}$ (episode) & 2e4 & 2e4 & 2e4 & 2e4 & N/A & N/A & 2e4 & 2e4 & 2e4 & 2e4 \\
        TD$(\lambda)$ & N/A & 0.6 & 0.6 & 0.8 & N/A & N/A & 0.6 & 0.8 & N/A & N/A \\
        Steps per train & N/A & N/A & N/A & N/A & 100 & 100 & N/A & N/A & N/A & N/A\\
        Target-net soft-update rate  & N/A & N/A & N/A & N/A & 5e-3 & 1e-1 & N/A & N/A & N/A & N/A\\
        Entropy loss weight & N/A & N/A & N/A & N/A & 1e-3 & 1e-2 & N/A & N/A & N/A & N/A\\
        Replay buffer size & N/A & 5000 & N/A & N/A & 1e4 & 1e4 & N/A & N/A & N/A & N/A \\ 
        Batch size (episode) & N/A & 32 & N/A & N/A & 2 & 8 & N/A & N/A & N/A & N/A  \\ 
        Num attention head & N/A & N/A & N/A & N/A & 4 & N/A & N/A & N/A & N/A & N/A  \\
    \bottomrule
    \end{tabular} 
\end{table}

\end{document}